\documentclass[twocolumn,final]{elsarticle}

\usepackage{graphicx}
\usepackage {subcaption}
\usepackage[ruled,linesnumbered]{algorithm2e}
\usepackage{amsmath}

\usepackage{ycviu}
\usepackage{framed,multirow}

\usepackage{amssymb}
\usepackage{latexsym}

\usepackage{url}
\usepackage{xcolor}
\definecolor{newcolor}{rgb}{.8,.349,.1}

\journal{Computer Vision and Image Understanding}

\begin{document}

\thispagestyle{empty}

\clearpage
\thispagestyle{empty}
\ifpreprint
  \vspace*{-1pc}
\fi

\begin{table*}[!th]
\ifpreprint\else\vspace*{-5pc}\fi

\section*{Graphical Abstract (Optional)}
To create your abstract, please type over the instructions in the
template box below.  Fonts or abstract dimensions should not be changed
or altered.

\vskip1pc
\fbox{
\begin{tabular}{p{0.9\textwidth}p{.5\textwidth}}
\bf Class Knowledge Overlay to Visual Feature Learning for Zero-Shot  Image Classification  \\
Cheng Xie, Ting Zeng, Hongxin Xiang, Keqin Li, Yun Yang, Qing Liu \\[1pc]
\includegraphics[width=0.85\textwidth]{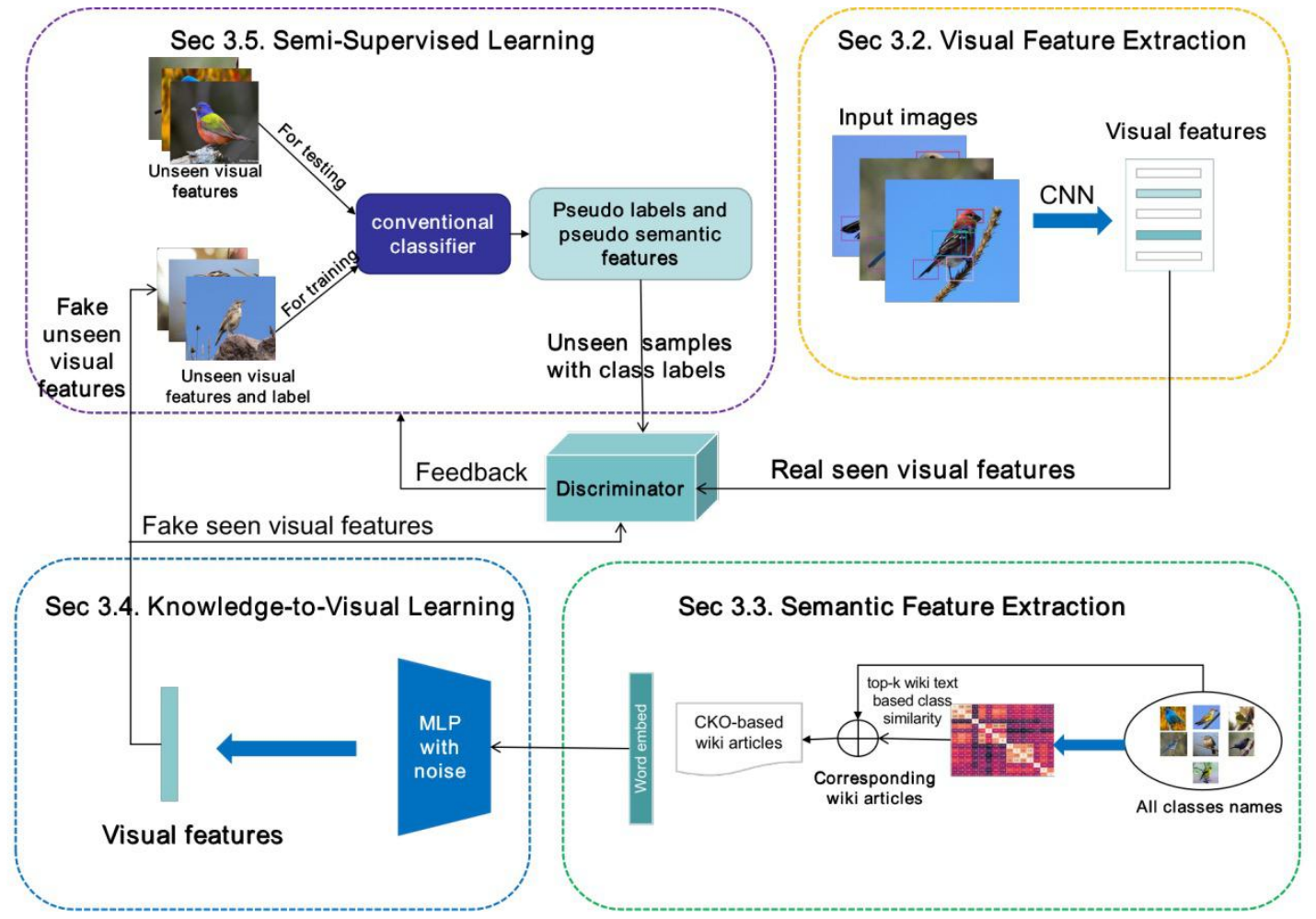}

New categories can be discovered by transforming semantic features into synthesized visual features without corresponding training samples in zero-shot image classification. Although significant progress has been made in generating high-quality synthesized visual features using generative adversarial networks, guaranteeing semantic consistency between the semantic features and visual features remains very challenging. In this paper, we propose a novel zero-shot learning approach, GAN-CST, based on class knowledge to visual feature learning to tackle the problem. The approach consists of three parts, class knowledge overlay, semi-supervised learning and triplet loss. It applies class knowledge overlay (CKO) to obtain knowledge not only from the corresponding class but also from other classes that have the knowledge overlay. It ensures that the knowledge-to-visual learning process has adequate information to generate synthesized visual features. The approach also applies a semi-supervised learning process to re-train knowledge-to-visual model. It contributes to reinforcing synthesized visual features generation as well as new category prediction. We tabulate results on a number of benchmark datasets demonstrating that the proposed model delivers superior performance over state-of-the-art approaches.
\end{tabular}
}

\end{table*}

\clearpage
\thispagestyle{empty}

\ifpreprint
  \vspace*{-1pc}
\else
\fi

\begin{table*}[!t]
\ifpreprint\else\vspace*{-15pc}\fi

\section*{Research Highlights (Required)}

\vskip1pc

\fboxsep=6pt
\fbox{
\begin{minipage}{.95\textwidth}
\vskip1pc
\begin{itemize}

  \item We propose a novel ZSL model based on knowledge-to-visual learning that outperforms state-of-the-art models on several benchmarks.
  \item We propose class knowledge overlay (CKO) to obtain more knowledge from similar categories that effectively improves the effect of knowledge-to-visual learning.
  \item We re-train the ZSL model by applying synthetic examples on a semi-supervised learning (SSL) process. It significantly reinforces category training and predicting.

\end{itemize}
\vskip1pc
\end{minipage}
}

\end{table*}

\clearpage

\ifpreprint
  \setcounter{page}{1}
\else
  \setcounter{page}{1}
\fi

\begin{frontmatter}

\title{Class Knowledge Overlay to Visual Feature Learning for Zero-Shot  Image Classification}

\author[1]{Cheng \snm{Xie}}
\author[1]{Ting \snm{Zeng}}
\author[1]{Hongxin \snm{Xiang}}
\author[2]{Keqin \snm{Li}}
\author[1]{Yun \snm{Yang}\corref{cor1}}
\cortext[cor1]{Yun Yang}
\ead{yangyun@ynu.edu.cn}
\author[1]{Qing \snm{Liu}}

\address[1]{Software Department, Yunnan University, Kunming 650091, China}
\address[2]{Department of Computer Science State University of New York New Paltz, New York, USA}

\received{1 May 2013}
\finalform{10 May 2013}
\accepted{13 May 2013}
\availableonline{15 May 2013}
\communicated{S. Sarkar}

\begin{abstract}
New categories can be discovered by transforming semantic features into synthesized visual features without corresponding training samples in zero-shot image classification. Although significant progress has been made in generating high-quality synthesized visual features using generative adversarial networks, guaranteeing semantic consistency between the semantic features and visual features remains very challenging. In this paper, we propose a novel zero-shot learning approach, GAN-CST, based on class knowledge to visual feature learning to tackle the problem. The approach consists of three parts, class knowledge overlay, semi-supervised learning and triplet loss. It applies class knowledge overlay (CKO) to obtain knowledge not only from the corresponding class but also from other classes that have the knowledge overlay. It ensures that the knowledge-to-visual learning process has adequate information to generate synthesized visual features. The approach also applies a semi-supervised learning process to re-train knowledge-to-visual model. It contributes to reinforcing synthesized visual features generation as well as new category prediction. We tabulate results on a number of benchmark datasets demonstrating that the proposed model delivers superior performance over state-of-the-art approaches.
\end{abstract}

\begin{keyword}
\MSC 41A05\sep 41A10\sep 65D05\sep 65D17
\KWD Image understanding \sep Zero-shot learning \sep Image classification \sep Knowledge representation \sep Generative adversarial network \sep Semi-supervised learning

\end{keyword}

\end{frontmatter}


\section{Introduction}
\label{sec1}
Humans can distinguish at least 30,000 basic object categories and any more subordinate ones \cite{Biederman1987Recognition}. Humans can also create new categories dynamically from a few or even zero examples \cite{Fu2017Recent}. In contrast, most existing computer vision techniques require tens of hundreds of labeled examples to learn a recognition model. Besides, it is difficult to guarantee that the recognition model is fully trained for each category, especially since many new categories do not have samples. Inspired by the humans' ability to recognize without seeing the samples, the research area of zero-shot learning (ZSL) has received increasing attention in recent years.

In ZSL, unseen categories without examples can be recognized by transferring knowledge obtained from the seen categories \cite{Long2017Zero}. Specifically, ZSL is designed to find an intermediate knowledge representation (e.g., attributes or textual features) to transfer the knowledge learned from the seen categories to the unseen ones \cite{wang2018zero}. There are three paradigms for transferring knowledge:

(1) To use the semantic attributes to annotate images while new categories can be predicted by extracting and compositing semantic attributes from new examples \cite{5206594}. However, the performance of these methods is rather primitive because the methods are proposed at the early stage of ZSL, which cannot find a proper way to integrate the attributes into the image.

(2) To use semantic embedding methods \cite{Akata_2013_CVPR,wang2018zero,ji2018stacked,2018Zero,Chen2019Structurally} to learn the mapping from visual space to the semantic space. Ref.\cite{wang2018zero} builds upon the recently introduced Graph Convolutional Network (GCN) \cite{Kipf2016Semi} and proposes an approach that uses both semantic embeddings and the categorical relationships to predict the classifiers. Ref.\cite{Chen2019Structurally} proposes a novel zero-shot learning  model that forms a neighborhood-preserving structure in the semantic embedding space and utilizes it to predict for unseen classes. Ref.\cite{2018Zero} proposes a novel manifold distance computed on a semantic class prototype graph, which takes into account the rich intrinsic semantic structure. Other semantic embedding algorithms have also been investigated such as semi-supervised max-margin learning frameworks \cite{2015Max,2016Semi} or multi-task learning \cite{hwang2011sharing,jayaraman2014decorrelating,hwang2014unified}. However, semantic embedding always suffers from the domain shift problem \cite{7053935} since the learning process is over-fitted with embedded attributes.

(3) To use knowledge-to-visual mapping technology to embed the attributes or Wikipedia articles into an image, Ref.\cite{Kodirov_2017_CVPR} reduces the domain shift and the overfitting problem, effectively. Most state-of-the-art ZSL models are based on knowledge-to-visual mapping \cite{huzero,Tao2017AttnGAN,wang2018zero,Zhu_2018_CVPR,2018Feature,2020Generative}, which can be classified into GAN-based methods \cite{Zhu_2018_CVPR,2018Feature,2020Generative} and VAE-based methods \cite{2017Zero,2018Generalized}. GAN-based methods use category semantics and Gaussian noise as inputs to the generator to generate visual features. The generator is trained to perform a minimum-maximum game with the discriminator. The VAE-based method associates the conditional generator network with an additional encoder that approximates the posterior distribution in order to infer the latent factors, and trains the two models by maximizing the lower limit of variation. However, to our best knowledge, the best result obtained from the state-of-the-art ZSL models has only 12.5\% accuracy in the Caltech UCSD Birds-2011(CUB) dataset (a common dataset widely used for ZSL task) \cite{ji2018stacked}. This value is slightly lower than that of the standard recognition tasks. Consequently, there are still many challenges to be overcome in ZSL models.

In this study, we identify two critical problems in the ZSL process that might affect its performance. The first problem is inadequate knowledge, which is caused by two aspects. On the one hand, the semantic features are not enough to describe the fine-grained visual features of a category; on the other hand, the semantic features and visual features are not fully expressed when embedded, especially in two very similar categories with no difference in embedding space. The second problem is inadequate examples. Because the seen classes may rarely (or almost never) intersect the unseen classes, it is difficult to achieve better performance only by relying on the seen class examples. Especially in the same attribute or text description, the visual appearance may be significantly different. For example, pigs and zebras have the same attribute "tail" semantically, but they are completely different visually.
In this study, to solve the first problem, we propose a class knowledge overlay calculation method to gather more knowledge from similar categories that help the model to learn more knowledge. To solve the second problem, we propose a semi-supervised process to generate synthetic examples to re-train the ZSL model that helps the model to predict unseen categories. The experimental results show that our approach outperforms the state-of-the-art methods in several benchmark datasets. Succinctly, we highlight the contributions of the work as following:
\begin{enumerate}
  \item We propose a novel ZSL model based on knowledge-to-visual learning that outperforms state-of-the-art models on several benchmarks.
  \item We propose class knowledge overlay (CKO) to obtain more knowledge from similar categories that effectively improves the effect of knowledge-to-visual learning.
  \item We re-train the ZSL model by applying synthetic examples on a semi-supervised learning (SSL) process. It significantly reinforces category training and predicting.
\end{enumerate}

\section{Related works}
A key idea of zero-shot learning is to find a appropriate embedding space that seen and unseen classes can share. There are three types of embedding in zero-shot learning approaches, which are to (a) map from the visual feature space to the semantic space \cite{Frome2013DeViSE,7053935, ZslCCSEICLR, Socher2013Zero,Zhang_2015_ICCV}, (b) or conversely \cite{Ba2015Predicting, Pambala2019Unified, Xian2017Zero,Zhu_2018_CVPR}, (c) or jointly map from both the visual and semantic space to common space \cite{yang2014unified,lei2015predicting,Akata_2015_CVPR}, respectively.

\subsection{Semantic Attributes}
Semantic attributes refer to express a class or an instance using attributes. ZSL uses attributes as side information and consists of two steps: 1) to train the seen classes: gain knowledge about attributes; 2) to inference the unseen classes: classify some unseen objects via known knowledge. This is the first and most basic method of ZSL. In 2009, a pioneering study on ZSL, Ref. \cite{5206594}, proposed direct attribute prediction(DAP) and indirect attribute prediction(IAP). They are the main forms of attribute-based learning which learns the attribute classifier first and then seeks the most promising unseen class. Ref. \cite{10.1007/978-3-642-15555-0_10} proposed an author-topic model to describe the attribute-specific distributions of image features. Ref. \cite{6974493} has proposed a weighted version of DAP based on the observation probability of the attributes. However, attribute-based learning ignores the associations between different attributes, and it is more accurate in predicting attributes than classes. Furthermore, attributes need a large number of experts to label, which is inefficient. On the contrary, our approach does not depend on any prior attributes.

\subsection{Semantic Embedding}
Semantic embedding is a text-to-vector technique that can be used for mapping the visual feature to semantic space. The semantic embedding-based learning is one of the most widely used methods \cite{Frome2013DeViSE,7053935, ZslCCSEICLR, Socher2013Zero,Zhang_2015_ICCV}. Attribute label embedding(ALE) \cite{Akata_2013_CVPR}, proposed a label embedding framework to solve the prediction of classes aiming at the attribute learning directly. It not only takes attribute as side information but also takes word vector and hierarchy label embedding(HLE) as side information. Besides, inspired by ALE, Ref. \cite{Akata_2015_CVPR} proposed structured joint embedding(SJE), a structured joint framework and used various side information to replace the era of artificial annotation attributes in ZSL tasks. In 2016, LatEm \cite{Xian_2016_CVPR}, a nonlinear model of SJE, was proposed. It has a stronger expressive ability and can be adapted to different types of samples. The semantic similarity embedding (SSE) \cite{Zhang_2015_ICCV} not only maintains semantic consistency but also ensures the accuracy of classification. The above studies directly transfer the visual feature space to the semantic space, which leads to the problem of the large semantic gap problem. In 2017, Ref. \cite{Kodirov_2017_CVPR} introduced semantic autoencoder(SAE), a bidirectional encoding and decoding method that significantly reduces the semantic gap. However, the experimental results of SAE are not optimistic because the feature space transferring technique cannot eliminate the semantic gap.

\subsection{Semantic-to-Visual Mapping}
Different from semantic embedding, semantic-to-visual mapping is designed to learn the mappings from semantic space to visual space. Currently, most approaches follow the idea of semantic-to-visual mapping \cite{Ba2015Predicting, Pambala2019Unified, Xian2017Zero} and lead a new era of ZSL. The Ref. \cite{Zhu_2018_CVPR} combined the generative adversarial network (GAN) and ZSL to transform the ZSL problem into an "imagination" problem. The method implements semantic-to-visual mapping using "imagining" visual features from semantic features. Other studies \cite{akata2016multi, Akata_2015_CVPR, lei2015predicting, romera2015embarrassingly, shigeto2015ridge, yang2014unified, zhang2017learning, Zhu_2018_CVPR} show that these approaches yielded optimistic results. However, these approaches cannot "imagine" the visual features of the unseen classes if the corresponding semantics have not appeared.
In this study, we use both semantics from one class and the "class knowledge overlay" to obtain more semantics from other similar classes. This approach significantly enriches the semantics for semantic-to-visual mapping.

\section{Methodology}
\label{approach}
\begin{figure*}
\setlength{\abovecaptionskip}{0cm}
\setlength{\belowcaptionskip}{0cm}
\centering
\includegraphics[width=16cm,height=8cm]{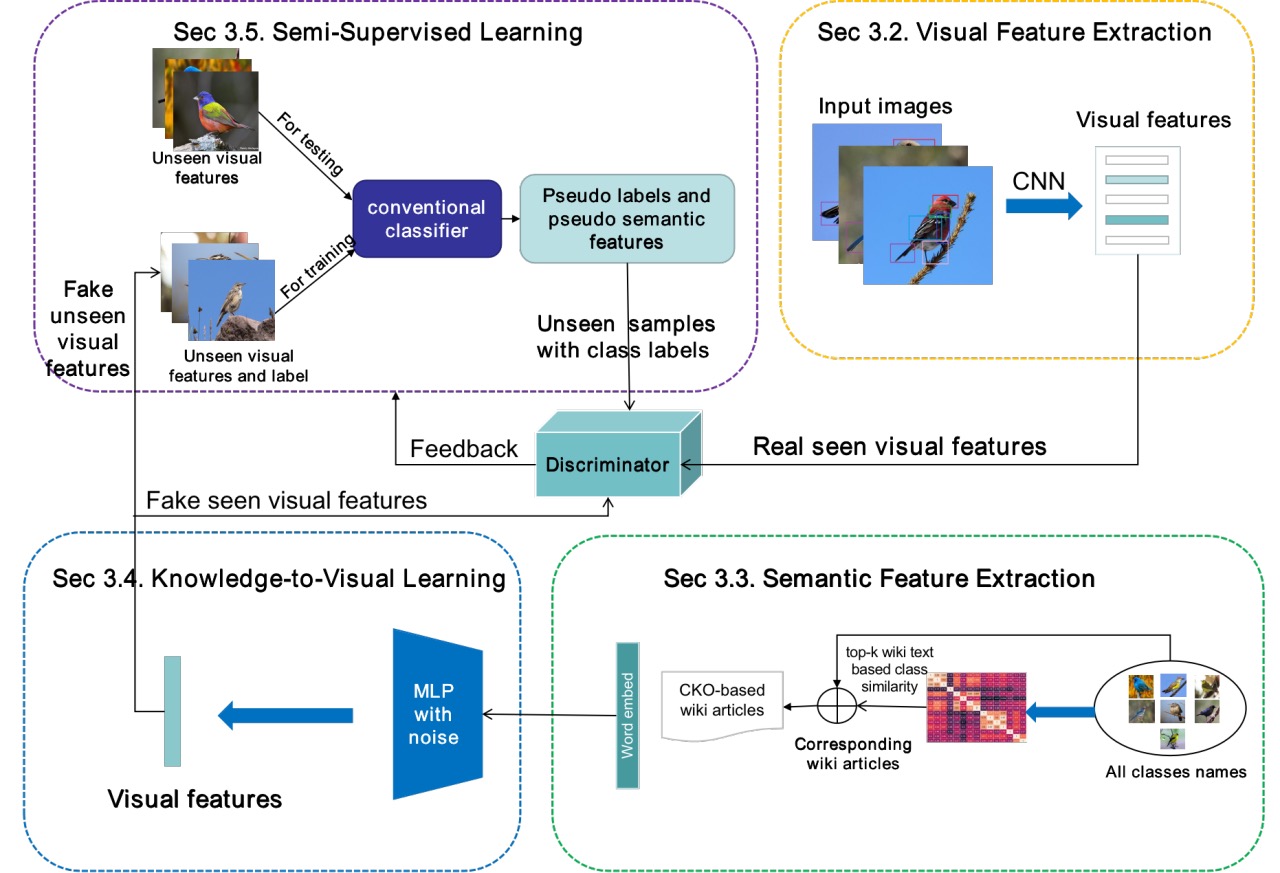}
\caption {Overall architecture: The approach first extracts semantic features by using the class knowledge overlay (CKO) (the green dotted part). In the module of knowledge-to-visual learning (the dotted blue part), multi-layer perceptrons (MLPs) with noise is used to transform class knowledge into synthesized visual features. The discriminator distinguishes the real and fake visual features and the correct classification labels. Finally, a semi-supervised learning mechanism (the dotted purple part) is used to iteratively select samples from unseen classes and their predicted pseudo-labels to augment the training data.}
\label{modelFrame}
\end{figure*}
The core of our approach is the design of a semantic-to-visual learning model. The overall method is demonstrated in Fig.\ref{modelFrame}. First, visual features are extracted by fast region-based convolutional network(fast-RCNN)(Section \ref{Visual Feature Extraction}). Then, semantic features are extracted from Wikipedia articles by CKO and TF-IDF (Section \ref{Semantic Feature Extraction}). Next, a GAN model is trained with triplet loss to "imagine" the synthetic visual features from semantic features (Section \ref{Semantic to Visual Generation}). Finally, a semi-supervised learning (SSL) algorithm is used to re-train the GAN model.

\subsection{Notations}
\label{Notations}
Suppose there is a series of data points $({d,y})$ from the original image dataset $D$ and label $Y$ respectively. We use subscripts $u$ and $s$ to represent datasets of unseen and seen classes after splitting the dataset, respectively. The visual features $x \in V$ can be extracted by using original images $d$. The semantics of seen and unseen categories are represented as $t_s,t_u$, which come from the semantic space $T$. For the $i$-th class, the representation of the class name is $E_i \in E$, where $E$ is the sets of all class names. The goal of ZSL is to predict $y_u$ based on $x_u$ and $t_u$. Generator $G$ and discriminator $D$ are represented as $R^T\times R^M\to R^V,R^V\to \{0,1\} \times L_{cls}$ where $R^M$ represents the mapping relationship of semantic features into visual features, and $L_{cls}$ represents the corresponding class labels in visual features $V$. We converted the parameters of $G$ and $D$ into $\theta$ and $ \omega $.


\subsection{Visual Feature Extraction}
\label{Visual Feature Extraction}
The visual features are extracted by the visual feature extraction methods described below: the fast-RCNN framework and the VGG16 architecture are used as the backbones to detect seven parts of the birds. First, the features of the input images $d \in D$ are extracted by VGG16. The proposed region of interest(ROI) pooling layer in \cite{Girshick2015Fast} is input into an n-ways softmax layer and a boundary box regression. Then, it is regarded as a detected visual feature when the proposed area is larger than a confidence threshold; otherwise, it is regarded as a missing part. Finally, the detected region is input into the visual encoder subnet and eventually encoded into 512-dimensional feature vectors for each part. The visual features of these seven parts are concatenated together to form 3584-dimensional visual features $x\in V$.

\subsection{Semantic Feature Extraction}
\label{Semantic Feature Extraction}

\begin{figure}
\
\includegraphics[width=8cm,height=3cm]{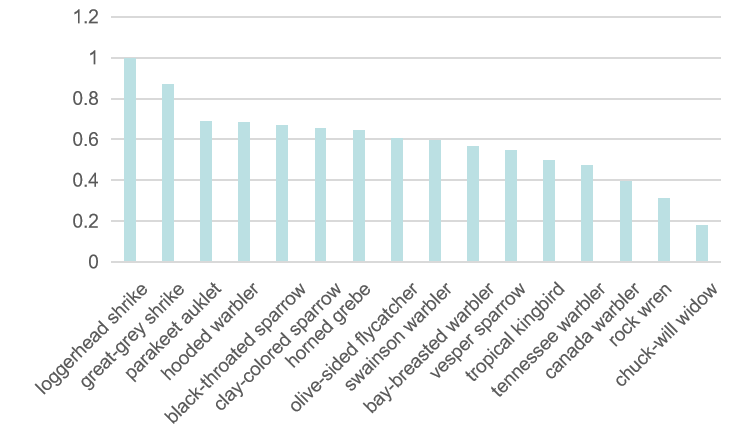}
\caption {The example of similarity scores of Logger-Head Shrike with other classes}
\label{word2vecSimWithShrike}
\end{figure}

\textbf{Class Knowledge Overlay:} The overall flow of the CKO algorithm is shown in the Algorithm \ref{CSOAlgorithm}. First, the word2vec is applied to transform each class to a vector. Second, a cosine similarity is used to calculate the similarity scores among the class vectors and the top-k similar classes of each class are ranked. Finally, the Wikipedia text of the category is represented by concatenating its own Wikipedia text and the Wikipedia text of the top-k similar classes. Fig.\ref{word2vecSimWithShrike} shows the similarity results of Logger-Head Shrike and other classes. Obviously, Logger-Head Shrike has a high similarity score to Great-Grey Shrike, which demonstrates that Great-Grey Shrike is likely to contain the knowledge of Logger-Head Shrike.\\
\indent \textbf{Semantic Embedding:} The Wikipedia texts are tokenized into words, firstly. Then, some necessary preprocesses, such as removing stop words, porter stemmer, and tokenization \cite{Porter2013An}, are applied to reduce inflected words to their word stem. Finally, the text encoder, TF-IDF, is used to extract and embed the semantic features.

\begin{algorithm}[t]
    \caption{CKO algorithm}
    \label{CSOAlgorithm}
    \KwIn{The number of unique labels $n$, similarity ranking $k$, text encoder $\Phi$,word2vec model $w2v$, class name of each class $E$, original Wikipedia articles of each class $A_o$ \;}
    \KwOut{class knowledge overlay between classes $SM \in R^{n \times n}$, overlay-based Wikipedia articles $A_s$, encoded text vectors $\Phi(A_s)$ \;}
    Initialize $SM \gets \emptyset, A_s \gets A_o $\;

      \For{$i=1;i \le n;i++$}
      {
      $s1 \gets w2v(E_i)$\;
      \For{$j=1;j\le n;j++$}
        {
        $s2 \gets w2v(E_j)$\;
        $SM[i][j] \gets distance(s1, s2)$\;
       }
       Rank $SM[i]$ in descending order and select the top $k$ classes, and mark them as $k_c$\;
      \For{$c=1;c \le k;c++$}
      {
      Add the wikipedia article of class $k_c$ to $A_s^i$\;
      }
      }
     $\Phi(A_s) \leftarrow$ encoding $A_s$ with TF-IDF encoder\;
    Return $SM$, $A_s$, $\Phi(A_s)$ \;
\end{algorithm}

\subsection{Knowledge-to-Visual Learning}
\label{Semantic to Visual Generation}
\textbf{Visual Feature Generation:} Text encoder $ \phi $  is used to embed texts. The embedded texts $ \phi ( T_{c} )$ are used as input to a generator ($G$ for short), which is a multi-layer perceptron with random noise $z$. Through this process, visual features $ \widetilde{x} $ can be generated by $ G_{ \theta } ( T_{c} ,z)$.

Because of the sparsity of training data (about 60 pictures per class of CUB datasets, and the distribution of visual features has about 3500 dimensions), it is difficult for the generator to achieve good results in transforming class knowledge into visual features. Ref.\cite{Zhu_2018_CVPR} reported that classes have the following characteristics in the visual space: the distance of intra-classes is short, the distance of inter-classes is long, and an overlap rarely occurs. Therefore, a new constraint can be added to the knowledge-to-visual features generation to make the synthetic visual features have the same visual distribution as the seen classes. The constraint is defined as follows:
\begin{equation}
\label{tripletLoss}
\begin{aligned}
L_{tripletloss} = & max \biggl( \frac{1}{C}\mathop{\sum}\limits_{{c=1}}^C \biggl(\frac{1}{n}\mathop{\sum}\limits_{{i=1}}^ndist(\widetilde{x}_s^c,x_s^{c,i})\\
&-\frac{1}{m}\mathop{\sum}\limits_{{i=1}}^mdist(\widetilde{x}_s^c,x_s^{\bar{c},i})\biggl)+margin, 0\biggl)
\end{aligned}
\end{equation}
where $C$ is the number of seen classes, $x_s^{c,i}$ is the $i$th visual feature of class $c$, $\widetilde{x}_s^c$ denotes the synthetic-visual features of category $C$ in the seen class, $\bar{c}$ denotes a class that does not belong to class $c$, $margin$ represents the minimum distance between two different class clusters, and $dist$ represents any measure. In this study, Euclidean distance is used as a measure. Finally, the loss of generator is defined as:
\begin{equation}
\label{tripletLoss}
\begin{aligned}
L_{G} = & \mathbb{E}[D_\omega(G_{\theta}(T,z))]-\mathbb{E}[D_\omega(x)] \\
        & +\frac{1}{2}(L_{cls}(G_\theta(T,z))+L_{cls}(x))+\lambda_t L_{tripletloss}
\end{aligned}
\end{equation}
where the first two terms approximate Wasserstein distance of the distribution of real features and fake features, the third and forth terms are classification losses of real and synthesized features. $\lambda_t$ is a regularization coefficient.

\textbf{Discriminator:} The discriminator ($D_\omega$ for short) accepts two inputs: fake visual features from $G$ or real visual features from images. Then it propagates them forward to a full connection layer with a ReLu activator. Next, two subnetworks are used to distinguish whether features are real or fake and classify the category label of these features. The loss function of $D_\omega$ is the same with the previous work\cite{Zhu_2018_CVPR}.

\subsection{Semi-supervised Learning}
\label{SemiSupervisedZSL}
During each SSL iteration, a conventional classifier are trained by using examples from $(G(t_u, z),y_u)$. In this paper, the conventional classifier is k-NearestNeighbor model. Then, the classifier predicts pseudo-labels, which have highest class probability in all classes, for each unseen class sample in $d_u$. Those samples whose class probability is above a certain threshold are stored in a set of $D_p=\{(D_s,\widetilde{y_u})\}$. In the next training, the training set $D_s$ is updated to $D_s\cup D_p$.
Because at the beginning of training, the model only trains the seen classes data. After the semi-supervised learning, the unseen classes with pseudo labels will be added to the training set. If the pseudo label is marked as unseen class, then a new class is introduced in the training set.
So we need to dynamically add new neurons to the subnetwork in the discriminator, which are used to classify new classes, and include this new category when calculating the triplet loss. The detailed training process of GAN with SSL is shown in Algorithm \ref{GANBasedOnSSL}.

\begin{algorithm}[!htb]
    \caption{Semi-supervised Learning for synthetic examples training}
    \label{GANBasedOnSSL}
    \KwIn{inter-class distance $margin$, confidence threshold $\psi$,the maximal loops $N_{step}$, early stopping coefficient $p$, the number of iterations for SSL $N_{ssl}$, the batch size $m$, Adam hyperparameters $\alpha$, $\beta_1$, $\beta_2$ \;}

    Initialize $p \gets 100, n_d \gets 5, \alpha \gets 0.001, \beta_1 \gets 0.5, \beta_2 \gets 0.9$ \;
      \For{$i=1;n \le N_{ssl};i++$}
      {
      \For{$step \gets 1;step \le N_{step} \&\& p_{count} \le p;step++$}
      {
      \For{$j=1;j \le 5;j++$}
      {
      Sample a minibatch of $m$ images $x$, matching texts $T$, random noise $z$ \;
      $\widetilde{x} \leftarrow G_\theta(T,z) $ \;
      Compute the discriminator loss $L_D$ \;
      $\omega\leftarrow Adam(\bigtriangledown_\omega[L_G], \theta, \alpha, \beta_1, \beta_2)$\;
      }
      Initialize each set in $\{Pos_{set}^c\}_{c=1}^C $ to $\emptyset$, $\{Neg_{set}^c\}_{c=1}^C $ to $\emptyset$, $L_{tripletloss}=0$ \;
      Sample a minibatch of $m$ class labels $c$, matching texts $T_c$, random noise z \;
      $\widetilde{x} \leftarrow G_\theta(T_c,z)$ \;
      Compute the generator loss $L_G$ \;
            \For{$j=1;j \le m;j++$}
	  {
	  Select  $n_1$ images $pos$ of the same classes as $c_j$ and $n_2$ images $neg$ of different classes from $c_j$ \;
	  $Pos_{set}^j \leftarrow pos, Neg_{set}^j \leftarrow neg $ \;
	  $L_{tripletloss}+=max(\frac{1}{n_1}\sum_t^{n_1}||\widetilde{x}_j - Pos_{set}^{j,t}||_2-\frac{1}{n_2}\sum_t^{n_2}||\widetilde{x}_j - Neg_{set}^{j,t}||_2, 0)+margin$
	  }
	   $\theta \leftarrow Adam(\bigtriangledown_\theta[L_G], \theta, \alpha, \beta_1, \beta_2)$\;
        Calculate the accuracy of seen classes and determine whether to stop early \;
      }

      Sample unseen images $x_u$, matching texts $T_{test}$ \;
      Train conventional classifier $Model$ using $x_u$ \;
      $y_{pro} \gets Model.predict(T_{test})$ \;

       add $x_u[y_{pro} \ge \psi]$ and corresponding texts to $x_s$ \;
	   Modify discriminator model structure \;
      }
\end{algorithm}
\subsection{Training and Testing}
\label{Training and Testing}
\textbf{Training:} Semantic features are extracted using the proposed class knowledge overlay(CKO), and visual features are extracted. through real images and generators. Then, ACGAN is trained with $n_{iter}$ iterations, including the training generator's ability to generate  visual features using semantic features with tripletloss, and the the training discriminator to judge visual features as fake or real and predict the class labels.  After the ACGAN training is completed, the generator uses the visual features generated by the semantic features of unseen classes and the corresponding semantic labels to train the traditional classifier(eg. Decision Tree, SVM,...). The trained classifier will give the label probability for visual features of the unseen class. For labels with a probability higher than a certain threshold, their visual features are added to the training set. Repeat the above process until the $n_{ssl}$ semi-supervised process is executed.

\textbf{Testing:} After training, we obtain the generation model $G$, which can transform semantic features of classes into synthetic visual features. In the testing process, the model compares the real visual features (from the new coming image) with the synthetic visual features (from the text of class). Then, the model decides the class of the new coming image.

\section{Experiments}

\subsection{Experimental Setup}
\subsubsection{Datasets}
\begin{figure}
\setlength{\abovecaptionskip}{0cm}
\setlength{\belowcaptionskip}{-0.5cm}
\includegraphics[width=8cm,height=5cm]{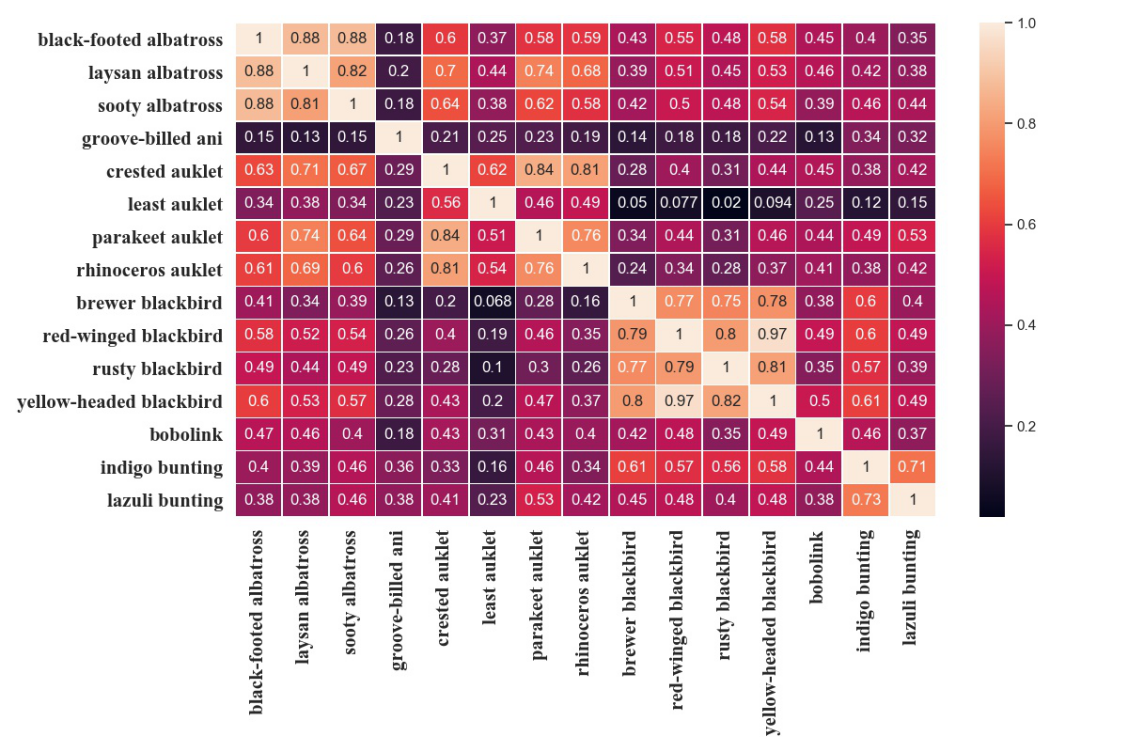}
\caption {An example of class knowledge overlay between different classes of CUB. The x-axis and y-axis represent classes selected from the CUB dataset, respectively. Each lattice represents the knowledge overlay of the corresponding x-axis and y-axis categories: the darker the color, the higher the knowledge overlay; the lighter the color, the lower the knowledge overlay.}
\label{similarity matrix}
\end{figure}

Our approach was compared with the state-of-the-art methods on two benchmarks: Caltech UCSD Birds-2011(CUB) and North America Birds(NAB). The CUB dataset contains 200 fine-grained classes of the birds with 11,788 images. The NAB dataset is a larger dataset of 48,562 images across 1011 bird classes. Besides, the raw textual sources from English Wikipedia-v01.02.2016 are adopted. Fig.\ref{similarity matrix} shows the class knowledge overlay of CUB data set. The class knowledge is embedded into vectors by using word2vec. The overlay is calculated by using Euclidean distance.
This obviously shows that CKO not only integrates the semantic features of the same parent category (such as black-footed albatross and laysan albatross, up to 88\% similarity), but also integrates the semantic features of different parent categories with high similarity (laysan albatross and parakeet auklet, up to 74\% similarity), and class overlay of category information of different superclass can add more semantic features.


\subsubsection{Split Methods}

In zero-shot learning, there are two commonly used training/testing set segmentation methods: Super-Category-Shared splitting (SCS) and Super-Category-Exclusive splitting (SCE), which are used in \cite{Elhoseiny2017Link,huzero,ji2018stacked,Zhu_2018_CVPR}. In the case of SCS-split, there are more than one seen class belonging to the same super category for each unseen class. For example, the classes "Tennessee Warbler" and "Wilson Warbler" are in the training set and in the testing set, respectively, but the super category is "Warbler" in CUB2011. Same as CUB2011, "Cooper's Hawk" in the training set and "Harris's Hawk" in the testing set have the same super category "Hawk" in NABirds. Compared with SCS, in the case of SCE-split, the classes with the same super category either belong to the training set(seen) or to the testing set(unseen). For instance, if "Caspian Tern" is selected as the training set, then all other terns are selected as the training set. Therefore, in SCE, the correlation between the seen and unseen classes is minimal. Consequently, the classification accuracy based SCE-split is lower than the SCS-split.
\subsubsection{Evaluation Metric}
In this study, top-1 accuracy is used to evaluate the predictive performance of the model. Top-1 accuracy has been widely used in existing works \cite{Elhoseiny2017Link, Li2018Discriminative, Meng2018Self, Sariyildiz_2019_CVPR, Xian2017Zero} to choose the highest prediction probability as the final result.
\subsubsection{Implementation Details}

\begin{table}
\begin{center}
\caption{The hyperparameters of the model under different settings. $margin$ denotes the distance between different classes of sample clusters in triplet loss; $topK$ denotes the overlay of K class texts; $confidence$ denotes the threshold for semi-supervised learning to select samples.}
\begin{tabular}{|l|l|l|l|l|}
\hline
                   & \multicolumn{2}{l|}{CUB} & \multicolumn{2}{l|}{NAB} \\ \hline
parameters            & SCS         & SCE        & SCS         & SCE        \\ \hline
margin      & 0.1        & 0.1          & 0.2           & 0.1          \\ \hline
topK & 4        & 1        & 3           & 1          \\ \hline
confidence & 0.5        & 0.7        & 0.6        & 0.4        \\ \hline
\end{tabular}
\label{superParamSelection}
\end{center}
\end{table}

\textbf{Semantic Features: }In this study, the Wikipedia text was used as side information to match some visual features with the words in it. Although Wikipedia texts are more expressive and discriminating than attribute representations, they usually have more noise. In this case, the methods described in Section \ref{Semantic Feature Extraction} were used to process the Wikipedia texts. Then, TF-IDF was used to extract the semantic features from the processed texts. The dimension of these features is 7551 and 13,217 in CUB2011 and NAB Wikipedia datasets, respectively.

\indent \textbf{Visual Features: }There are seven parts of the input image, (1) head, (2) back, (3) belly, (4) breast, (5) leg, (6) wing, and (7) tail for capturing the different characteristics of birds. For each part of the bird, a 512-dimensional vector can be obtained after applying the Multi-Layer Perceptron(MLP) with two hidden layers(each with a size of 512). For the CUB2011 dataset, seven bird parts were used as visual features, whereas in the NAB dataset, the "leg" part was deleted since there are no annotations for the "leg" part in the NAB dataset. The remaining six parts were retained as visual features. Therefore, the feature dimensions extracted from CUB2011 and NAB datasets were 3584 and 3072, respectively.\\

\indent \textbf{Model Setting: } The seen dataset is divided into training set and validation set according to the ratio of 9:1. The semantic features were input into a MLP in the semantic-to-visual generation method. Firstly, the MLP used a 1000-dimensional full connection layer to reduce the dimensions of the semantic features. Then random noise was added to the semantic features of dimension reduction. Finally, two full-connection layers with LeakyRelu and Tanh were used to generate visual features. Table \ref{superParamSelection} shows the hyperparameters of our method under different settings. For the study of hyperparameters see Section \ref{superParamStudy}. Our model is trained with Adam, using the default parameters $\beta_1=0.9,\beta_2=0.999$, and the learning rate $ \eta=0.001 $. And $MaxIter=10000$ and $batchsize=1000$ are set. The KNN model (K = 20) was trained to evaluate the seen class and the unseen class in every 40 iterations. The unseen class accuracy, which the highest generalized accuracy of the seen class in validation set corresponds to, was selected as the final result. The generalized accuracy is calculated as follows:
\begin{equation}
\label{generizedAcc}
G_{acc}=\frac{1}{m}\sum_{\lambda}^{J}\frac{1}{N}\sum_{n=0}^{N}{I} \biggl(\mathop{\arg\max} \biggl(\sum_{i=n_{tcls}}^{n_{cls}}\hat{y}_n^i+\lambda \biggl) ==y_n \biggl)
\end{equation}
where, $m=\frac{J-\lambda}{\delta}$, the $\delta$ presents the update frequency of $\lambda$. $N$ denotes the sample numbers of a seen classes, $n_{cls}$ denotes the number of all classes, $n_{tcls}$ denotes the seen classes number, $\hat{y}_n^i$ denotes the prediction probability of the nth sample on the $i$ class, $y_n$ denotes the real class label of the nth sample. Argmax function indicates the predictive label of $\hat{y}$. In this study, we set $\lambda=-2, J=2$ and $\delta=0.01$.
\subsection{Performance Evaluation}

\subsubsection{Comparative Methods}
Nine latest methods were used in the comparisons with our methods: ZSLNS \cite{Qiao2016Less}, SynCfast \cite{Changpinyo2016Synthesized}, ZSLPP \cite{Elhoseiny2017Link}, GDAN \cite{Huang_2019_CVPR}, CIZSL \cite{2019arXiv190401109E}, CANZSL \cite{chen2020canzsl}, GAN-ZSL \cite{Zhu_2018_CVPR}, CorrectionNet \cite{huzero}, S$^2$GA-DET \cite{ji2018stacked}. All the comparisons used the same splits. For the first three methods, we cite the results from \cite{Zhu_2018_CVPR}. The results of last five methods are cited in their respective papers, which report the maximum of the results. For GDAN, we reproduce report the best results by using source code it provide.

The performance of our method (GAN-CST) was evaluated on two benchmark datasets by using two segmentation methods: SCE and SCS.
As shown in Table \ref{top1AccWithOtherModels}, compared to the state-of-the-art methods, GAN-CST obtain the best result in SCS-split on CUB dataset and SCE-split on NAB dataset, which increases by 0.66\% and 7.22\%. Compared with the latest generative ZSL methods (CIZSL, CANZSL, GDAN), except for the slightly lower SCE-split of CUB dataset, 14.1\% vs. 14.4\%, our method exceeds these methods by up to 11.83\%. Since there are some correlations between the training set and the test set in the SCS-split, it is difficult to detect more correlations by adding some test samples into the training dataset with semi-supervised learning. Therefore, the improvement of GAN-CST is not apparent on the SCS-split. However, some improvements were still achieved compared to the GAN-ZSL method in the SCS-split.


\newcommand{\tabincell}[2]{\begin{tabular}{@{}#1@{}}#2\end{tabular}}
\begin{table}[!htb]
\centering
\setlength{\abovecaptionskip}{0.cm}
\setlength{\belowcaptionskip}{-0.cm}
\caption{Top-1 accuracy (\%) on CUB and NAB datasets with two split settings.}
\begin{tabular}{|p{3.5cm}|p{0.8cm}|p{0.8cm}|p{0.8cm}|p{0.8cm}|}
\hline
                   & \multicolumn{2}{l|}{CUB} & \multicolumn{2}{l|}{NAB} \\ \hline
Methods            & SCS         & SCE        & SCS         & SCE        \\ \hline
ZSLNS \cite{Qiao2016Less}     & 29.1        & 7.3        & 24.5        & 6.8        \\ \hline
SynCfast \cite{Changpinyo2016Synthesized}   & 28.0        & 8.6        & 18.4        & 3.8        \\ \hline
ZSLPP \cite{Elhoseiny2017Link}     & 37.2        & 9.7        & 30.3        & 8.1        \\ \hline
GAN-ZSL \cite{Zhu_2018_CVPR}    & 43.7        & 10.3       & 35.6        & 8.6        \\ \hline
CorrectionNet \cite{huzero}    & 45.8        & 10.0       & 37.0        & 9.5        \\ \hline
S$^2$GA-DET \cite{ji2018stacked}    & 42.9        & 10.9       & \textbf{39.4}        & 9.7        \\ \hline
CIZSL \cite{2019arXiv190401109E}     & 44.6        & \textbf{14.4}        & 36.6        & 9.3        \\ \hline
CANZSL \cite{chen2020canzsl}     & 45.8        & 14.3        & 38.1        & 8.9        \\ \hline
GDAN \cite{Huang_2019_CVPR}     & 44.2        & 13.7        & 38.3        & 8.7        \\ \hline
GAN-CST               & \textbf{46.1}      &14.1  &38.6        & \textbf{10.4}       \\ \hline
\end{tabular}
\label{top1AccWithOtherModels}
\end{table}

\subsubsection{Ablation Study}
\begin{table}[!htb]
\caption{Ablation Study. The top-1 accuracy of different combinations is tabulated. TL, CKO, and SSL represent the triplet loss, class knowledge overlay, and semi-supervised learning, respectively.}
\centering
\begin{tabular}{|l|l|l|l|l|}
\hline
\multicolumn{1}{|c|}{}                   & \multicolumn{2}{c|}{CUB}                                    & \multicolumn{2}{l|}{NAB} \\ \hline
\multicolumn{1}{|c|}{method}             & \multicolumn{1}{c|}{SCS}     & \multicolumn{1}{c|}{SCE}     & SCS         & SCE        \\ \hline
\multicolumn{1}{|c|}{ACGAN} & \multicolumn{1}{c|}{43.7} & \multicolumn{1}{c|}{10.3} & 35.6            & 8.6           \\ \hline
\multicolumn{1}{|c|}{ACGAN (+TL)} & \multicolumn{1}{c|}{44.1} & \multicolumn{1}{c|}{11.6} & 35.9            & 8.9           \\ \hline
\multicolumn{1}{|c|}{ACGAN (+CKO)} & 44.6    & 12.1     & 37.3  & 9.3           \\ \hline
\multicolumn{1}{|c|}{ACGAN (+SSL)} & 43.8  & 10.9  & 36.6 & 8.8      \\ \hline
\multicolumn{1}{|c|}{ACGAN (+CKO+SSL)} &44.9 &11.6 & 36.2  & 9.2           \\ \hline
\multicolumn{1}{|c|}{ACGAN (+CKO+TL)} &44.3  & 13.1  & 38.1 & 8.1           \\ \hline
\multicolumn{1}{|c|}{ACGAN (+SSL+TL)} &44.6 &13.3 & 36.5 & 9.7       \\ \hline
\multicolumn{1}{|c|}{GAN-CST} & \textbf{46.1} & \textbf{14.1} & \textbf{38.6} & \textbf{10.4}  \\ \hline
\end{tabular}
\label{ablationStudy}
\end{table}
Extensive ablation experiments were conducted to observe the effect of triplet loss(TL), class knowledge overlay(CKO), semi-supervised learning(SSL) and their combinations on the results. Table \ref{ablationStudy} illustrates the results of the ablation studies. Note that ACGAN is our basic structure. Obviously, our method after adding each component exceeds ACGAN, which shows the effectiveness of each of our components. In addition, the table also shows that the combination of multiple components can improve the performance of the model in most cases. Therefore, the superposition of the methods has a positive correlation with the final prediction accuracy.

\subsection{Generalized Zero-shot Learning}

\begin{table}[!htb]
\centering
\setlength{\abovecaptionskip}{0.cm}
\setlength{\belowcaptionskip}{-0.cm}
\caption{AUSUC (\%) on CUB and NAB datasets with two split settings.}
\begin{tabular}{|p{3.5cm}|p{0.8cm}|p{0.8cm}|p{0.8cm}|p{0.8cm}|}
\hline
                   & \multicolumn{2}{l|}{CUB} & \multicolumn{2}{l|}{NAB} \\ \hline
Methods            & SCS         & SCE        & SCS         & SCE        \\ \hline
ZSLNS \cite{Qiao2016Less}     & 14.7        & 4.4        & 9.3        & 2.3        \\ \hline
SynCfast \cite{Changpinyo2016Synthesized}   & 13.1        & 4.0        & 2.7        & 3.5        \\ \hline
ZSLPP \cite{Elhoseiny2017Link}     & 30.4        & 6.1        & 12.6        & 3.5        \\ \hline
GAN-ZSL \cite{Zhu_2018_CVPR}    & 35.4        & 8.7       & 20.4        & 5.8        \\ \hline
CorrectionNet \cite{huzero}    & \textbf{41.9}      & 9.0       & 25.4        & 7.6        \\ \hline
CIZSL \cite{2019arXiv190401109E}     & 39.2        & 11.9        & 24.5        & 6.4        \\ \hline
CANZSL \cite{chen2020canzsl}     & 40.2        & 12.5        & \textbf{25.6}        & 6.8        \\ \hline
GDAN \cite{Huang_2019_CVPR}     & 38.7        & 10.9        & 24.1        & 5.9        \\ \hline
GAN-CST               & 40.5      &\textbf{12.7}  &24.9        & \textbf{7.9}       \\ \hline
\end{tabular}
\label{AUSUCTable}
\end{table}

In the ZSL domain, it is not sufficient to only consider the performance of the unseen classes. A more generalized evaluation criterion is needed. In \cite{Zhu_2018_CVPR}, a generalized evaluation metric, which considers the accuracy of the seen and unseen classes, was proposed for ZSL. A balance parameter was used to draw the curves of the seen and unseen classes(SUC, the accuracy of the seen classes is the vertical axis and the accuracy of the unseen classes is the horizontal axis), and the area under SUC (AUSUC) was used to represent the generalization ability of the ZSL model.
Table \ref{AUSUCTable} shows the AUSUC scores between our method and the other methods. The AUSUC score of our method increased by 1.6\% and 3.95\%, respectively, on two benchmark datasets with SCE splitting compared to the other methods. In the SCS-split, our method is slightly lower than CorrectionNet and CANZSL, only 1.4\% and 0.7\%, but still surpasses a large number of the state-of-the-art methods.

\begin{figure}[!htb]
\centering
    \begin{subfigure}[b]{0.22\textwidth}
        \includegraphics[height=2.5cm,width=4cm]{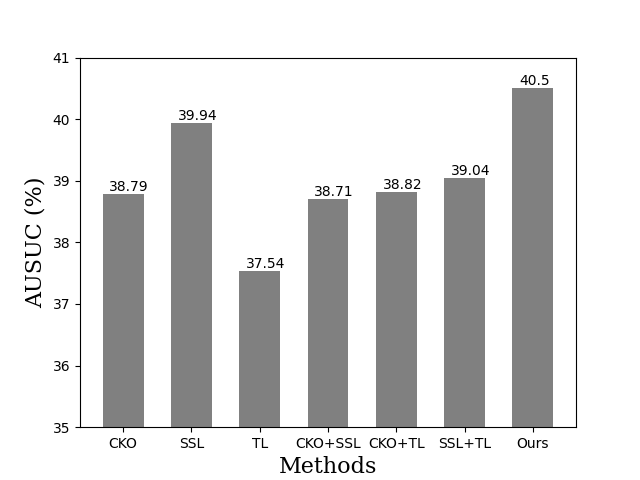}
        \caption{CUB with SCS splitting}
        \label{pic1}
    \end{subfigure}
    \begin{subfigure}[b]{0.22\textwidth}
        \includegraphics[height=2.5cm,width=4cm]{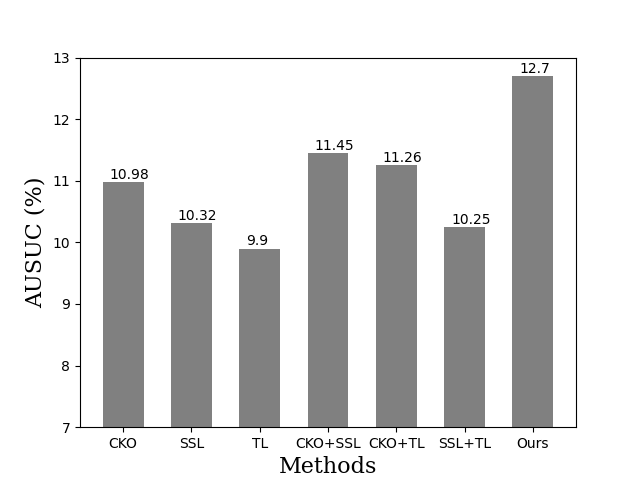}
        \caption{CUB with SCE splitting}
        \label{pic1}
    \end{subfigure}
    \begin{subfigure}[b]{0.22\textwidth}
        \includegraphics[height=2.5cm,width=4cm]{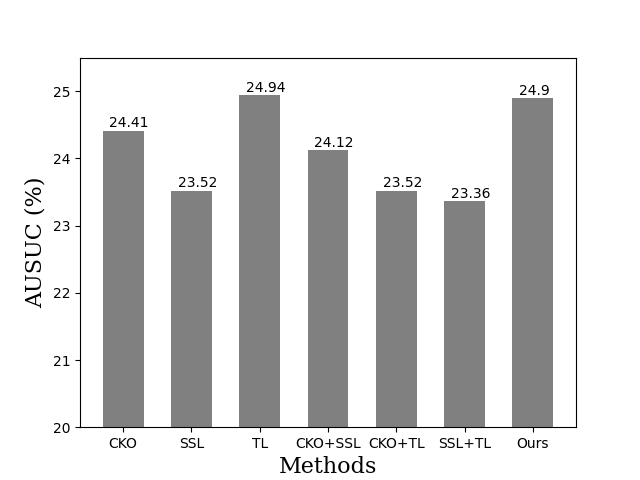}
        \caption{NAB with SCS splitting}
        \label{pic1}
    \end{subfigure}
    \begin{subfigure}[b]{0.22\textwidth}
        \includegraphics[height=2.5cm,width=4cm]{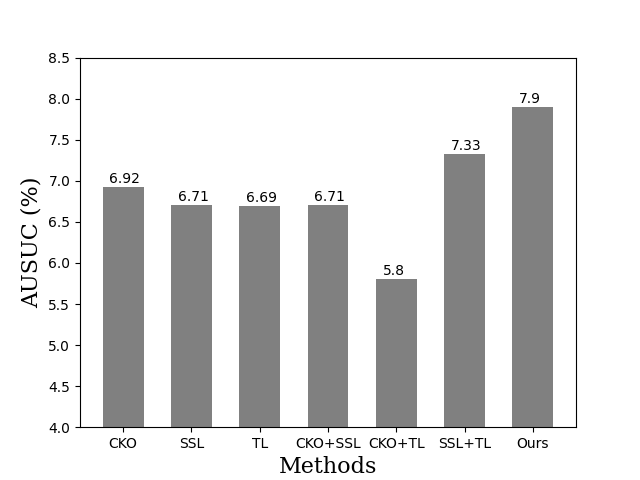}
        \caption{NAB with SCE splitting}
        \label{pic1}
    \end{subfigure}
\caption{AUSUC scores of our approach on two benchmark datasets with two split settings. TL, CKO, and SSL represent the triplet loss, class knowledge overlay, and semi-supervised learning, respectively.}
\label{AUSUCFigOurMethod}
\end{figure}

We also evaluate the AUSUC scores of each component in our method. Fig.\ref {AUSUCFigOurMethod} shows that the effect of triplet loss on the result performance is relatively stable, while the performance of CKO and SSL methods changes greatly. This is because the CKO and SSL sometimes introduce some noise that affects the training of the model. However, the generalization of each combination reached the state-of-the-art standard.

In addition, we use another GZSL setting that emerges recently to evaluate the proposed method on AwA1 and AwA2 datasets. These two datasets are based on attribute and respectively contain 30,475 and 37,322 images of 200 animals with 40 seen and 10 unseen classes with 85-dimensional attributes. In this setting, test set includes data samples from both the seen and unseen classes. We follow the same setting in \cite{Xian2017Zero}, which adopt the average per-class top-1 accuracy S and U, as well as their harmonic mean to evaluate the performance of the model and combines the seen and unseen classes as the search space. The Table \ref{GBUResults} shows that the proposed method compared with seven latest methods. The results show our GAN-CST exceeds a large number of the latest method. Especially in the S of AwA1 dataset and S and H of AwA2 dataset, the best performances are achieved, which are 97.2\%, 94.0\% and 85.6\%, respectively. An obvious rule can be observed: our GAN-CST improves the U accuracy while ensuring a high S accuracy. Although AwA1 has achieved better performance on U and H compared to our method, the performance of our method on S far exceeds it, and the accuracy on S is almost 100\%. This shows that our method can well retain the discriminative features of seen classes while improving the performance of unseen classes.
\begin{table}[!htb]
\centering
\setlength{\abovecaptionskip}{0.cm}
\setlength{\belowcaptionskip}{-0.cm}
\caption{Comparative results (\%) of state-of-the-arts with the same setting used in \cite{Xian2017Zero}. U and S are the Top-1 accuracies tested on unseen classes and seen classes, respectively. H is the harmonic mean of U and S. The hyperparameters are margin=200, k=1 and confidence=0.9.}
\begin{tabular}{|p{2.7cm}|p{0.55cm}p{0.55cm}p{0.55cm}|p{0.55cm}p{0.55cm}p{0.55cm}|} \hline
& \multicolumn{3}{c|}{AwA1} & \multicolumn{3}{c|}{AwA2} \\ \hline
Methods  &S  &U  &H  &S  &U  &H   \\ \hline
f-CLSWGAN \cite{Qiao2016Less}    &61.4  &57.9  &59.6  &68.9  &52.1  &59.4  \\ \hline
CADA-VAE \cite{Changpinyo2016Synthesized}    &72.8  &57.3  &64.1  &75.0  &55.8  &63.9  \\ \hline
LisGAN \cite{Elhoseiny2017Link}    &76.3  &52.6  &62.3  &-  &-  &-  \\ \hline
GMN \cite{Zhu_2018_CVPR}    &79.2  &70.8  &74.8  &-  &-  &-        \\ \hline
GXE \cite{huzero}    &89.0  &\textbf{87.7}  &\textbf{88.4}  &90.0  &\textbf{80.2}  &84.8  \\ \hline
CE \cite{2019arXiv190401109E}    &87.7  &71.2  &78.6  &86.1  &71.3  &78.0        \\ \hline
Deep-CDM \cite{chen2020canzsl}    &-  &-  &-  &82.5  &77.6  &80.0        \\ \hline
GAN-CST   &\textbf{97.2}  &73.9  &84.0  &\textbf{94.0}  &78.6  &\textbf{85.6}        \\ \hline
\end{tabular}
\label{GBUResults}
\end{table}

\subsection{Zero-Shot Retrieval}

The task of zero-shot retrieval means to retrieve the relevant images from unseen classes giving the semantic representation of the specified class in unseen class set. We use mean average precision (mAP) to evaluate the performance. For comparing with other methods fairly, we report the performance of different settings in Table \ref{zslretrieval}: retrieving 25\%, 50\%, 100\% of the number of images for each class from the whole dataset are ranked based on their final semantic similarity scores. The precision is defined as the ratio of the number of correct retrieved images to that of all retrieved images.

Table \ref{zslretrieval} presents the comparison results of different approaches for mean accuracy precision (mAP) on CUB and NABird datasets. We note that the proposed approach has achieved consistent improvement compared with GAN-ZSL and beats all the competitors.

\begin{table}[!htb]
\centering
\caption{Zero-shot retrieval mAP (in \%) comparison on CUB and NAB datasets. The results of all the competitors are cited from \cite{Zhu_2018_CVPR}.}
\begin{tabular}{p{2.3cm}|p{0.58cm}|p{0.58cm}|p{0.58cm}|p{0.58cm}|p{0.58cm}|p{0.58cm}}
\hline
 &  \multicolumn{3}{c|}{CUB}       &  \multicolumn{3}{c}{NAB}     \\ \hline
Methods & 25 & 50 & 100 & 25 & 50 & 100 \\ \hline
ESZSL\cite{Romera2015An} &27.9 &27.3&22.7 &28.9 &27.8 &20.9   \\ \hline
ZSLNS\cite{Qiao2016Less} & 29.2 &29.5  &23.9   &28.8  &27.3  &22.1   \\ \hline
ZSLPP\cite{Elhoseiny2017Link} &42.3  &42.0  &36.6  &36.9 &35.7  &31.3   \\ \hline
GAN-Only\cite{Zhu_2018_CVPR} &18.0  &17.5  &15.2  &21.7  &20.3  &16.6  \\ \hline
GAN-ZSL\cite{Zhu_2018_CVPR} &49.7   &48.3   &40.3   &41.6   &37.8  &31.0  \\ \hline
GAN-CST&\textbf{51.6}& \textbf{50.4}& \textbf{43.6} &\textbf{44.9}  &\textbf{41.3}  &\textbf{35.0}  \\ \hline

\end{tabular}
\label{zslretrieval}
\end{table}

\begin{figure}[!htb]
\vspace{-0.45cm}
\setlength{\abovecaptionskip}{-0.7cm}  
\setlength{\belowcaptionskip}{-0cm}   
\centering
\includegraphics[width=8cm,height=6cm]{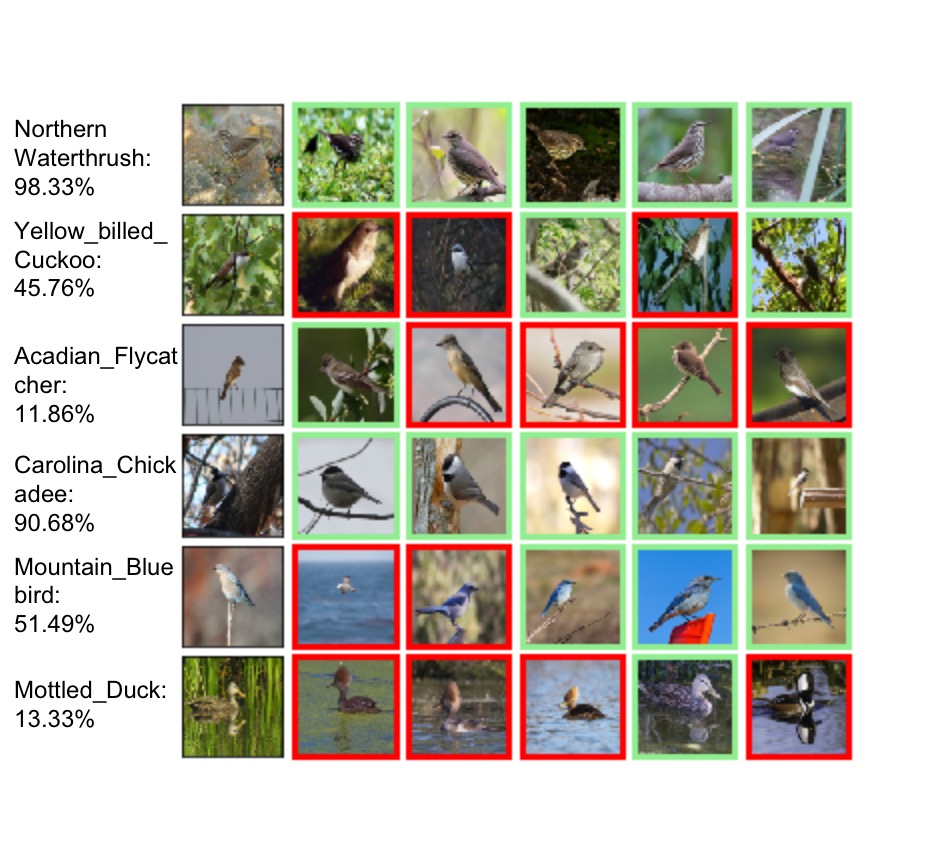}
\caption {Visualization samples of zero-shot retrieval with our approach. The first three rows are classes
from CUB dataset, and the rest are classes from NABird dataset. Correct and incorrect retrieved
instances are marked in lightgreen and red, respectively.}

\label{retrievalresult}
\end{figure}

We also visualize some qualitative results of our approach on two datasets, shown in Fig. \ref{retrievalresult}. Each row is a class, and the class name and precision are shown on the left. The first column is the benchmark. The following five columns are Top-5 without considering the instances in the first column. Some instances are hard to distinguish even for humans, but the model can recognize. For example, the top-5 retrieval images of class "Northern Waterthrush" are all from their ground truth class since their visual features are similar. However, the query "Mountain Bluebird" retrieves some instances from its affinal class "Florida Scrub Jay" since their visual features are too similar to distinguish.
\subsection{Hyperparameter Study}
\label{superParamStudy}
\begin{figure}[!htb]
\vspace{-0.4cm}
\setlength{\abovecaptionskip}{-0cm}  
\setlength{\belowcaptionskip}{-0cm}   
\centering
    \begin{subfigure}[b]{0.5\textwidth}
        \includegraphics[height=2.3cm,width=2.8cm]{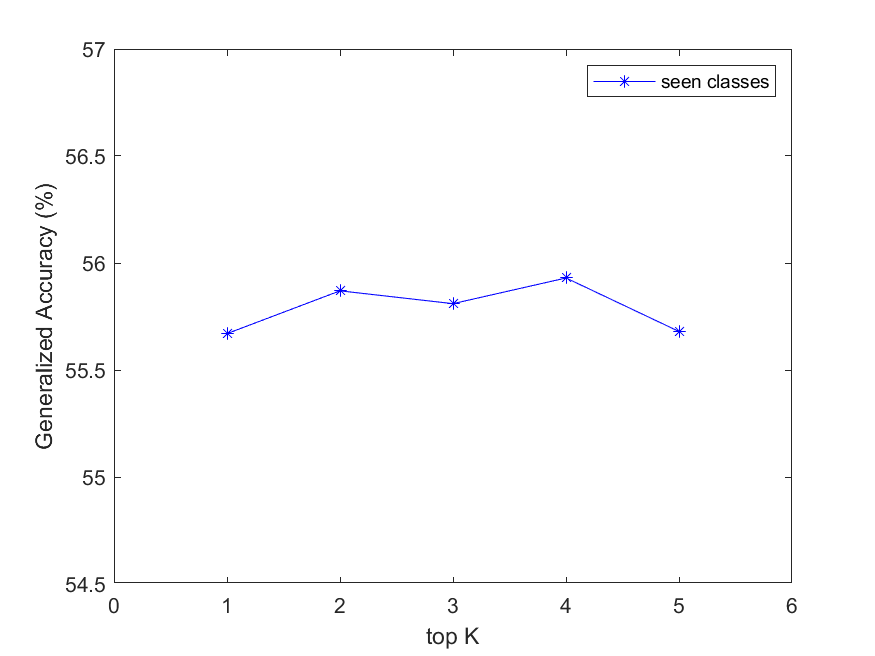}
        \includegraphics[height=2.3cm,width=2.8cm]{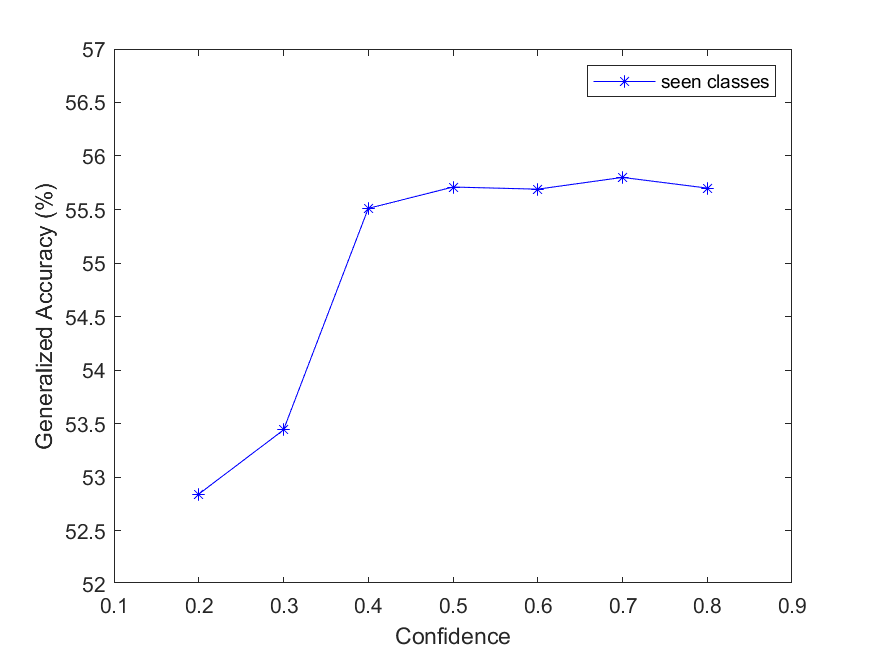}
        \includegraphics[height=2.3cm,width=2.8cm]{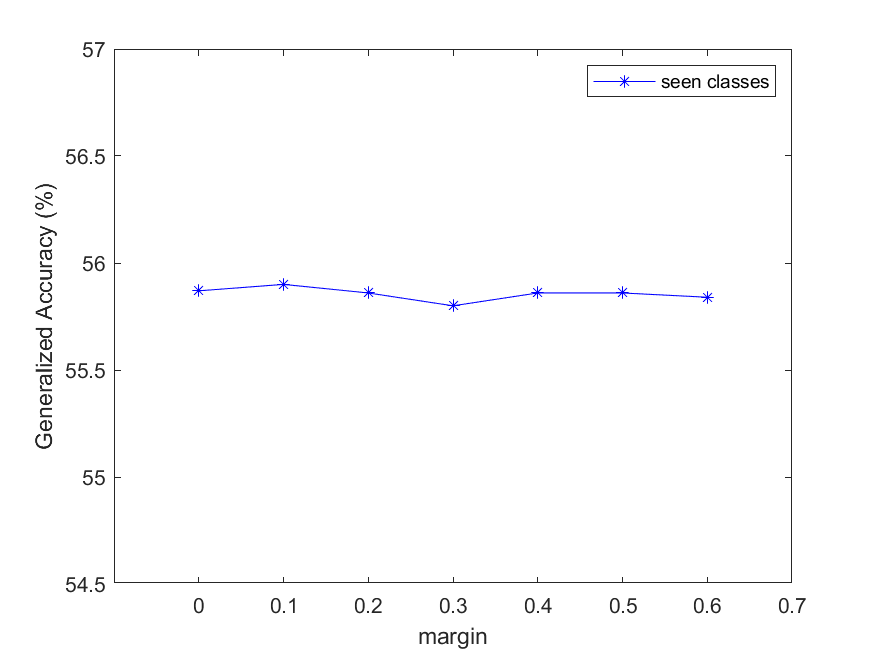}
        \caption{CUB Dataset}
        \label{pic1}
    \end{subfigure}
    \begin{subfigure}[b]{0.5\textwidth}
        \includegraphics[height=2.3cm,width=2.8cm]{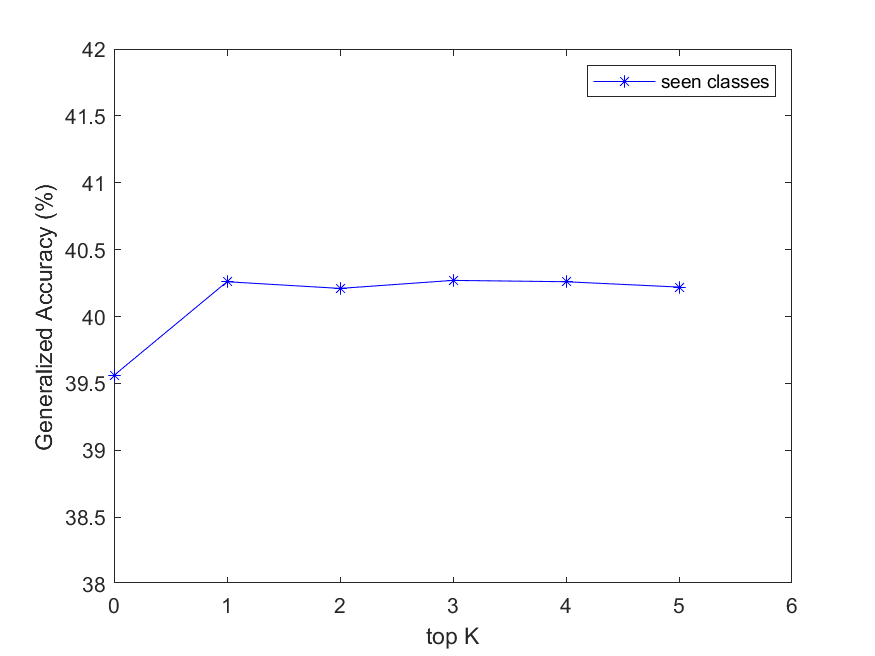}
        \includegraphics[height=2.3cm,width=2.8cm]{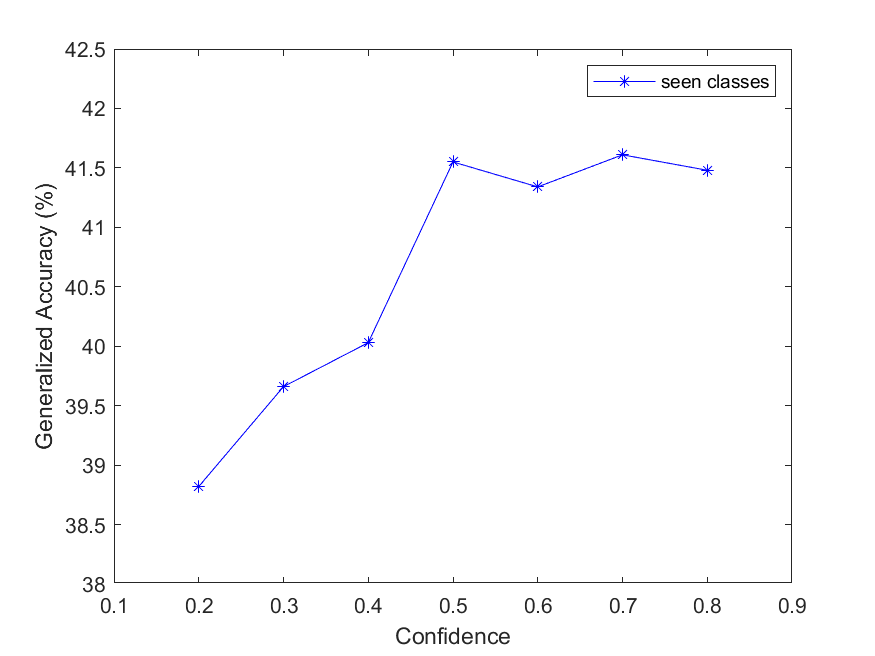}
        \includegraphics[height=2.3cm,width=2.8cm]{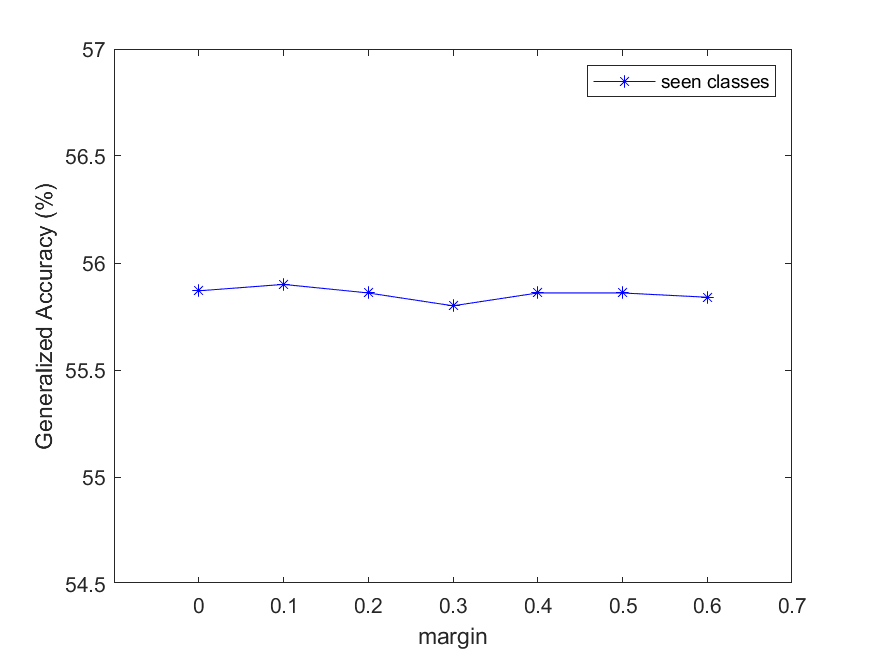}
        \caption{NAB Dataset}
        \label{pic1}
    \end{subfigure}
\caption{Generalized Accuracy on two benchmark datasets with SCS-split}
\label{superParamStudySCS}
\end{figure}
\begin{figure}[!htb]
\centering
\vspace{0cm}
\setlength{\abovecaptionskip}{-0cm}  
\setlength{\belowcaptionskip}{-0cm}   
    \begin{subfigure}[b]{0.5\textwidth} 
        \includegraphics[height=2.3cm,width=2.8cm]{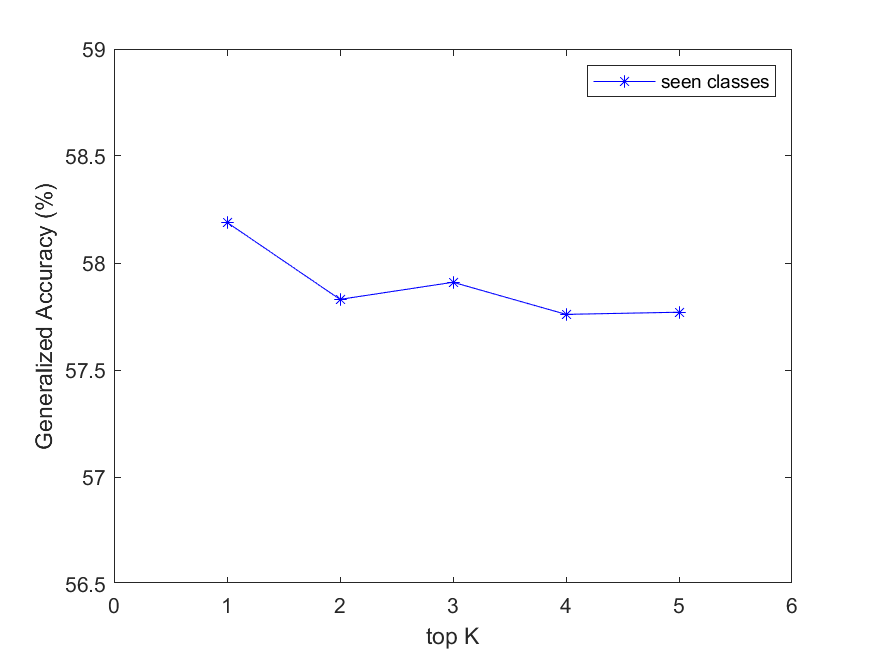}
        \includegraphics[height=2.3cm,width=2.8cm]{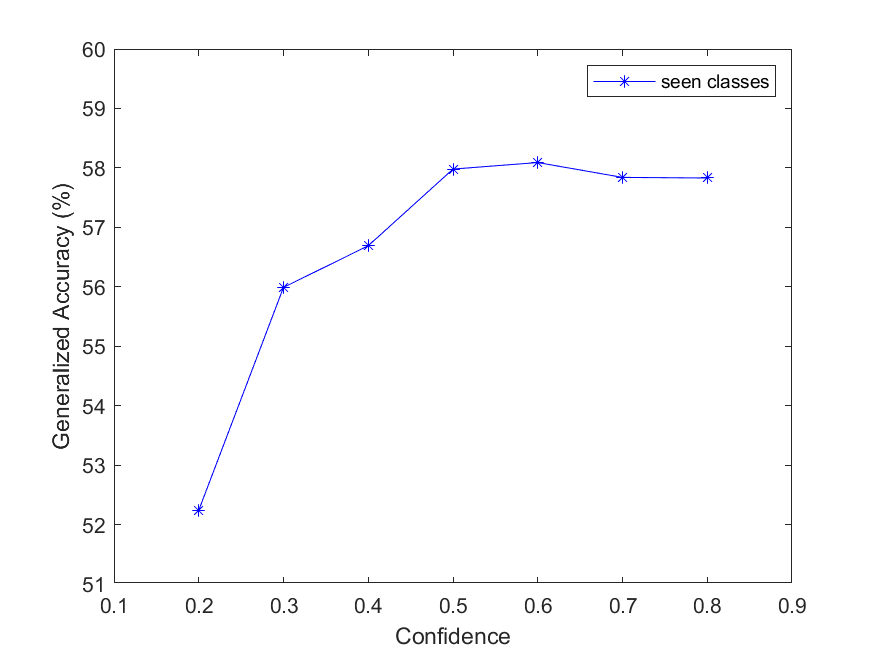}
        \includegraphics[height=2.3cm,width=2.8cm]{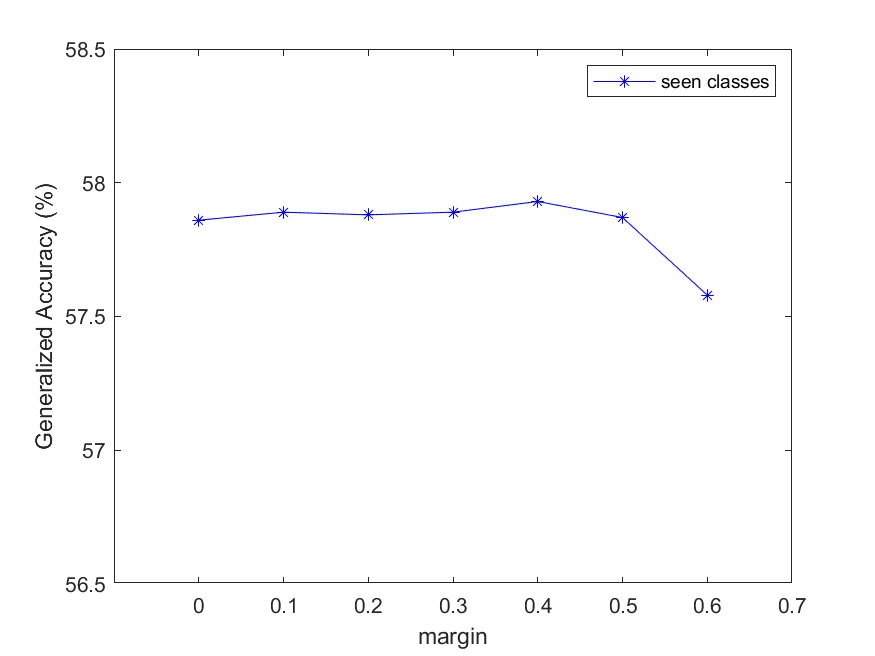}
        \caption{CUB Dataset}
        \label{pic1}
    \end{subfigure}
    \begin{subfigure}[b]{0.5\textwidth}
        \includegraphics[height=2.3cm,width=2.8cm]{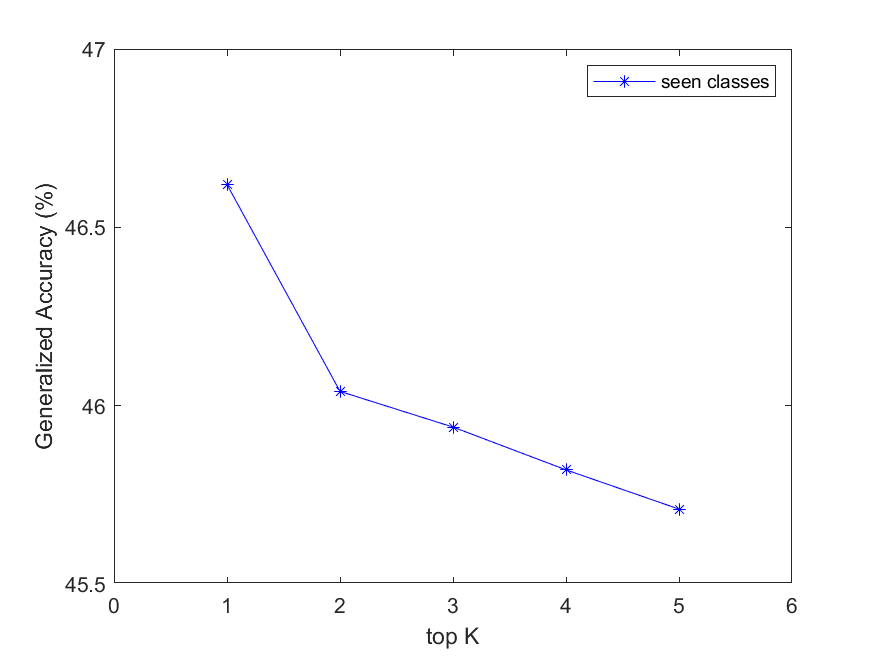}
        \includegraphics[height=2.3cm,width=2.8cm]{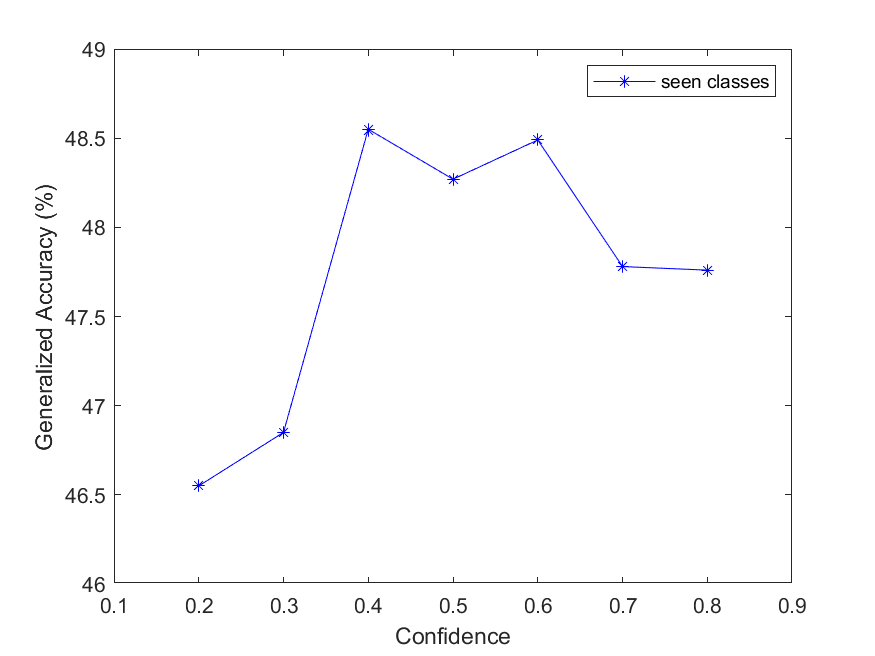}
        \includegraphics[height=2.3cm,width=2.8cm]{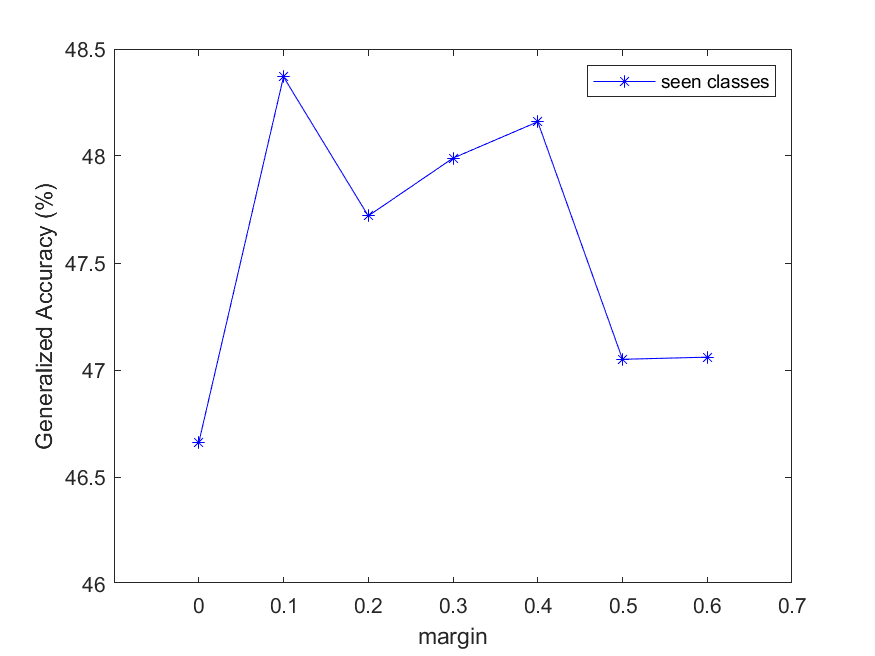}
        \caption{NAB Dataset}
        \label{pic1}
    \end{subfigure}
\caption{Generalized Accuracy on two benchmark datasets with SCE-split}
\label{superParamStudySCE}
\end{figure}

In this section, different hyperparameters were set to observe their impact on the performance of our model. Three groups of experiments were conducted. The hyperparameter settings are shown in Table \ref{superParamSetting}. Fig.\ref{superParamStudySCS} and \ref{superParamStudySCE} show the generalized accuracy curves with different splitting methods and different hyperparameters in two benchmark datasets. The horizontal axis represents the values of the hyperparameters, while the vertical axis represents the generalization accuracy of the seen classes (calculated by formula \ref{generizedAcc}). The values of the hyperparameters corresponding to the highest generalization accuracy are set as the parameters of the model. Table \ref{superParamSelection} summarizes the values of the hyperparameters in all groups. Our method is more robust than the other methods because of the gaps in the accuracy of the unseen classes with different hyperparameter settings.

\begin{table}[!htb]
\centering
\caption{Hyperparameters settings. TL, CKO and SSL represent triplet loss, class knowledge overlay and semi-supervised learning respectively.}
\begin{tabular}{|p{0.05cm}|p{1.3cm}|p{1.6cm}|p{4.5cm}|p|}
\hline
  & Methods               & Parameters & Candidate Values \\ \hline
1 & TL  & margin & \begin{tabular}[c]{@{}l@{}}0 to 0.8 with an interval of 0.1\end{tabular} \\ \hline
2 & CKO        & k & 1, 2, 3, 4, 5                                                                    \\ \hline
3 & SSL & confidence & \begin{tabular}[c]{@{}l@{}}0.2, 0.3, 0.4, 0.5, 0.6, 0.7, 0.8\end{tabular}  \\ \hline
\end{tabular}
\label{superParamSetting}
\end{table}

\section{Conclusion}
In this study, we developed a novel approach to solve the challenging zero-shot learning tasks. Our approach uses an ACGAN to transform semantic features into visual features. Meanwhile, class knowledge overlay and semi-supervised learning were used to solve the problem of the semantic consistency between the semantic features and visual features, respectively. Furthermore, triplet loss was introduced to expand the inter-class distances and shorten the intra-class distances. Extensive experiments showed that our approach significantly outperforms the state-of-the-art models on multiple zero-shot tasks. Our future works may focus on: 1) applying a sophisticated visual feature generation method to improve the quality of synthesized visual part; 2) instead of text embedding, knowledge graph embedding would be applied to enhance the ability of semantic representation.

\section*{Acknowledgments}
The authors wish to acknowledge the financial support from the Natural Science Foundation of China (NSFC) under Grant No.61876166 and 61663046 and the Young Researcher Promotion Project of China Association for Science and Technology under Grant W8193209.

\bibliographystyle{model2-names}
\bibliography{refs}

\begin{thebibliography}{53}
\expandafter\ifx\csname natexlab\endcsname\relax\def\natexlab#1{#1}\fi
\providecommand{\url}[1]{\texttt{#1}}
\providecommand{\href}[2]{#2}
\providecommand{\path}[1]{#1}
\providecommand{\DOIprefix}{doi:}
\providecommand{\ArXivprefix}{arXiv:}
\providecommand{\URLprefix}{URL: }
\providecommand{\Pubmedprefix}{pmid:}
\providecommand{\doi}[1]{\href{http://dx.doi.org/#1}{\path{#1}}}
\providecommand{\Pubmed}[1]{\href{pmid:#1}{\path{#1}}}
\providecommand{\bibinfo}[2]{#2}
\ifx\xfnm\relax \def\xfnm[#1]{\unskip,\space#1}\fi
\bibitem[{Akata et~al.(2016)Akata, Malinowski, Fritz and
  Schiele}]{akata2016multi}
\bibinfo{author}{Akata, Z.}, \bibinfo{author}{Malinowski, M.},
  \bibinfo{author}{Fritz, M.}, \bibinfo{author}{Schiele, B.},
  \bibinfo{year}{2016}.
\newblock \bibinfo{title}{Multi-cue zero-shot learning with strong
  supervision}, in: \bibinfo{booktitle}{Proceedings of the IEEE Conference on
  Computer Vision and Pattern Recognition}, pp. \bibinfo{pages}{59--68}.
\bibitem[{Akata et~al.(2013)Akata, Perronnin, Harchaoui and
  Schmid}]{Akata_2013_CVPR}
\bibinfo{author}{Akata, Z.}, \bibinfo{author}{Perronnin, F.},
  \bibinfo{author}{Harchaoui, Z.}, \bibinfo{author}{Schmid, C.},
  \bibinfo{year}{2013}.
\newblock \bibinfo{title}{Label-embedding for attribute-based classification},
  in: \bibinfo{booktitle}{The IEEE Conference on Computer Vision and Pattern
  Recognition (CVPR)}.
\bibitem[{Akata et~al.(2015)Akata, Reed, Walter, Lee and
  Schiele}]{Akata_2015_CVPR}
\bibinfo{author}{Akata, Z.}, \bibinfo{author}{Reed, S.},
  \bibinfo{author}{Walter, D.}, \bibinfo{author}{Lee, H.},
  \bibinfo{author}{Schiele, B.}, \bibinfo{year}{2015}.
\newblock \bibinfo{title}{Evaluation of output embeddings for fine-grained
  image classification}, in: \bibinfo{booktitle}{The IEEE Conference on
  Computer Vision and Pattern Recognition (CVPR)}.
\bibitem[{Ba et~al.(2015)Ba, Swersky, Fidler and
  Salakhutdinov}]{Ba2015Predicting}
\bibinfo{author}{Ba, J.L.}, \bibinfo{author}{Swersky, K.},
  \bibinfo{author}{Fidler, S.}, \bibinfo{author}{Salakhutdinov, R.},
  \bibinfo{year}{2015}.
\newblock \bibinfo{title}{Predicting deep zero-shot convolutional neural
  networks using textual descriptions} .
\bibitem[{Biederman(1987)}]{Biederman1987Recognition}
\bibinfo{author}{Biederman, I., .}, \bibinfo{year}{1987}.
\newblock \bibinfo{title}{Recognition-by-components: a theory of human image
  understanding}.
\newblock \bibinfo{journal}{Psychological Review} \bibinfo{volume}{94},
  \bibinfo{pages}{115--47}.
\bibitem[{Changpinyo et~al.(2016)Changpinyo, Chao, Gong and
  Sha}]{Changpinyo2016Synthesized}
\bibinfo{author}{Changpinyo, S.}, \bibinfo{author}{Chao, W.L.},
  \bibinfo{author}{Gong, B.}, \bibinfo{author}{Sha, F.}, \bibinfo{year}{2016}.
\newblock \bibinfo{title}{Synthesized classifiers for zero-shot learning}, in:
  \bibinfo{booktitle}{Computer Vision \& Pattern Recognition}.
\bibitem[{Chen et~al.(2019)Chen, Xiong, Gao and Xiong}]{Chen2019Structurally}
\bibinfo{author}{Chen, Y.}, \bibinfo{author}{Xiong, Y.}, \bibinfo{author}{Gao,
  X.}, \bibinfo{author}{Xiong, H.}, \bibinfo{year}{2019}.
\newblock \bibinfo{title}{Structurally constrained correlation transfer for
  zero-shot learning}, in: \bibinfo{booktitle}{2018 IEEE Visual Communications
  and Image Processing (VCIP)}.
\bibitem[{Chen et~al.(2020)Chen, Li, Luo, Huang and Yang}]{chen2020canzsl}
\bibinfo{author}{Chen, Z.}, \bibinfo{author}{Li, J.}, \bibinfo{author}{Luo,
  Y.}, \bibinfo{author}{Huang, Z.}, \bibinfo{author}{Yang, Y.},
  \bibinfo{year}{2020}.
\newblock \bibinfo{title}{Canzsl: Cycle-consistent adversarial networks for
  zero-shot learning from natural language}, in: \bibinfo{booktitle}{The IEEE
  Winter Conference on Applications of Computer Vision}, pp.
  \bibinfo{pages}{874--883}.
\bibitem[{Elhoseiny and Elfeki(2019)}]{2019arXiv190401109E}
\bibinfo{author}{Elhoseiny, M.}, \bibinfo{author}{Elfeki, M.},
  \bibinfo{year}{2019}.
\newblock \bibinfo{title}{Creativity inspired zero-shot learning}.
\newblock \bibinfo{journal}{2019 IEEE/CVF International Conference on Computer
  Vision (ICCV)} , \bibinfo{pages}{5783--5792}.
\bibitem[{Elhoseiny et~al.(2017)Elhoseiny, Zhu, Han and
  Elgammal}]{Elhoseiny2017Link}
\bibinfo{author}{Elhoseiny, M.}, \bibinfo{author}{Zhu, Y.},
  \bibinfo{author}{Han, Z.}, \bibinfo{author}{Elgammal, A.},
  \bibinfo{year}{2017}.
\newblock \bibinfo{title}{Link the head to the "beak": Zero shot learning from
  noisy text description at part precision}, in: \bibinfo{booktitle}{Computer
  Vision \& Pattern Recognition}.
\bibitem[{Frome et~al.(2013)Frome, Corrado, Shlens, Bengio, Dean, Ranzato and
  Mikolov}]{Frome2013DeViSE}
\bibinfo{author}{Frome, A.}, \bibinfo{author}{Corrado, G.S.},
  \bibinfo{author}{Shlens, J.}, \bibinfo{author}{Bengio, S.},
  \bibinfo{author}{Dean, J.}, \bibinfo{author}{Ranzato, M.},
  \bibinfo{author}{Mikolov, T.}, \bibinfo{year}{2013}.
\newblock \bibinfo{title}{Devise: A deep visual-semantic embedding model}, in:
  \bibinfo{booktitle}{International Conference on Neural Information Processing
  Systems}.
\bibitem[{{Fu} et~al.(2015){Fu}, {Hospedales}, {Xiang} and {Gong}}]{7053935}
\bibinfo{author}{{Fu}, Y.}, \bibinfo{author}{{Hospedales}, T.M.},
  \bibinfo{author}{{Xiang}, T.}, \bibinfo{author}{{Gong}, S.},
  \bibinfo{year}{2015}.
\newblock \bibinfo{title}{Transductive multi-view zero-shot learning}.
\newblock \bibinfo{journal}{IEEE Transactions on Pattern Analysis and Machine
  Intelligence} \bibinfo{volume}{37}, \bibinfo{pages}{2332--2345}.
\newblock \DOIprefix\doi{10.1109/TPAMI.2015.2408354}.
\bibitem[{Fu et~al.(2017)Fu, Tao, Jiang, Xue and Gong}]{Fu2017Recent}
\bibinfo{author}{Fu, Y.}, \bibinfo{author}{Tao, X.}, \bibinfo{author}{Jiang,
  Y.G.}, \bibinfo{author}{Xue, X.}, \bibinfo{author}{Gong, S.},
  \bibinfo{year}{2017}.
\newblock \bibinfo{title}{Recent advances in zero-shot recognition} .
\bibitem[{Fu et~al.(2018)Fu, Xiang, Kodirov and Gong}]{2018Zero}
\bibinfo{author}{Fu, Z.}, \bibinfo{author}{Xiang, T.},
  \bibinfo{author}{Kodirov, E.}, \bibinfo{author}{Gong, S.},
  \bibinfo{year}{2018}.
\newblock \bibinfo{title}{Zero-shot learning on semantic class prototype
  graph}.
\newblock \bibinfo{journal}{IEEE Transactions on Pattern Analysis and Machine
  Intelligence} \bibinfo{volume}{40}, \bibinfo{pages}{2009--2022}.
\bibitem[{Girshick(2015)}]{Girshick2015Fast}
\bibinfo{author}{Girshick, R.}, \bibinfo{year}{2015}.
\newblock \bibinfo{title}{Fast r-cnn}.
\newblock \bibinfo{journal}{Computer Science} .
\bibitem[{Hu et~al.()Hu, Xiong and Socher}]{huzero}
\bibinfo{author}{Hu, R.L.}, \bibinfo{author}{Xiong, C.},
  \bibinfo{author}{Socher, R.}, .
\newblock \bibinfo{title}{Zero-shot image classification guided by natural
  language descriptions of classes: A meta-learning approach} .
\bibitem[{Huang et~al.(2019)Huang, Wang, Yu and Wang}]{Huang_2019_CVPR}
\bibinfo{author}{Huang, H.}, \bibinfo{author}{Wang, C.}, \bibinfo{author}{Yu,
  P.S.}, \bibinfo{author}{Wang, C.D.}, \bibinfo{year}{2019}.
\newblock \bibinfo{title}{Generative dual adversarial network for generalized
  zero-shot learning}, in: \bibinfo{booktitle}{Proceedings of the IEEE/CVF
  Conference on Computer Vision and Pattern Recognition (CVPR)}.
\bibitem[{Huang et~al.(2020)Huang, Wang, Yu and Wang}]{2020Generative}
\bibinfo{author}{Huang, H.}, \bibinfo{author}{Wang, C.}, \bibinfo{author}{Yu,
  P.S.}, \bibinfo{author}{Wang, C.D.}, \bibinfo{year}{2020}.
\newblock \bibinfo{title}{Generative dual adversarial network for generalized
  zero-shot learning}, in: \bibinfo{booktitle}{2019 IEEE/CVF Conference on
  Computer Vision and Pattern Recognition (CVPR)}.
\bibitem[{Hwang et~al.(2011)Hwang, Sha and Grauman}]{hwang2011sharing}
\bibinfo{author}{Hwang, S.J.}, \bibinfo{author}{Sha, F.},
  \bibinfo{author}{Grauman, K.}, \bibinfo{year}{2011}.
\newblock \bibinfo{title}{Sharing features between objects and their
  attributes}, in: \bibinfo{booktitle}{CVPR 2011},
  \bibinfo{organization}{IEEE}. pp. \bibinfo{pages}{1761--1768}.
\bibitem[{Hwang and Sigal(2014)}]{hwang2014unified}
\bibinfo{author}{Hwang, S.J.}, \bibinfo{author}{Sigal, L.},
  \bibinfo{year}{2014}.
\newblock \bibinfo{title}{A unified semantic embedding: Relating taxonomies and
  attributes}, in: \bibinfo{booktitle}{Advances in Neural Information
  Processing Systems}, pp. \bibinfo{pages}{271--279}.
\bibitem[{Jayaraman et~al.(2014)Jayaraman, Sha and
  Grauman}]{jayaraman2014decorrelating}
\bibinfo{author}{Jayaraman, D.}, \bibinfo{author}{Sha, F.},
  \bibinfo{author}{Grauman, K.}, \bibinfo{year}{2014}.
\newblock \bibinfo{title}{Decorrelating semantic visual attributes by resisting
  the urge to share}, in: \bibinfo{booktitle}{Proceedings of the IEEE
  Conference on Computer Vision and Pattern Recognition}, pp.
  \bibinfo{pages}{1629--1636}.
\bibitem[{Ji et~al.(2018)Ji, Fu, Guo, Pang, Zhang et~al.}]{ji2018stacked}
\bibinfo{author}{Ji, Z.}, \bibinfo{author}{Fu, Y.}, \bibinfo{author}{Guo, J.},
  \bibinfo{author}{Pang, Y.}, \bibinfo{author}{Zhang, Z.M.}, et~al.,
  \bibinfo{year}{2018}.
\newblock \bibinfo{title}{Stacked semantics-guided attention model for
  fine-grained zero-shot learning}, in: \bibinfo{booktitle}{Advances in Neural
  Information Processing Systems}, pp. \bibinfo{pages}{5995--6004}.
\bibitem[{Kipf and Welling(2016)}]{Kipf2016Semi}
\bibinfo{author}{Kipf, T.N.}, \bibinfo{author}{Welling, M.},
  \bibinfo{year}{2016}.
\newblock \bibinfo{title}{Semi-supervised classification with graph
  convolutional networks} .
\bibitem[{Kodirov et~al.(2017)Kodirov, Xiang and Gong}]{Kodirov_2017_CVPR}
\bibinfo{author}{Kodirov, E.}, \bibinfo{author}{Xiang, T.},
  \bibinfo{author}{Gong, S.}, \bibinfo{year}{2017}.
\newblock \bibinfo{title}{Semantic autoencoder for zero-shot learning}, in:
  \bibinfo{booktitle}{The IEEE Conference on Computer Vision and Pattern
  Recognition (CVPR)}.
\bibitem[{{Lampert} et~al.(2009){Lampert}, {Nickisch} and
  {Harmeling}}]{5206594}
\bibinfo{author}{{Lampert}, C.H.}, \bibinfo{author}{{Nickisch}, H.},
  \bibinfo{author}{{Harmeling}, S.}, \bibinfo{year}{2009}.
\newblock \bibinfo{title}{Learning to detect unseen object classes by
  between-class attribute transfer}, in: \bibinfo{booktitle}{2009 IEEE
  Conference on Computer Vision and Pattern Recognition}, pp.
  \bibinfo{pages}{951--958}.
\newblock \DOIprefix\doi{10.1109/CVPR.2009.5206594}.
\bibitem[{Lei~Ba et~al.(2015)Lei~Ba, Swersky, Fidler
  et~al.}]{lei2015predicting}
\bibinfo{author}{Lei~Ba, J.}, \bibinfo{author}{Swersky, K.},
  \bibinfo{author}{Fidler, S.}, et~al., \bibinfo{year}{2015}.
\newblock \bibinfo{title}{Predicting deep zero-shot convolutional neural
  networks using textual descriptions}, in: \bibinfo{booktitle}{Proceedings of
  the IEEE International Conference on Computer Vision}, pp.
  \bibinfo{pages}{4247--4255}.
\bibitem[{Li and Guo(2015)}]{2015Max}
\bibinfo{author}{Li, X.}, \bibinfo{author}{Guo, Y.}, \bibinfo{year}{2015}.
\newblock \bibinfo{title}{Max-margin zero-shot learning for multi-class
  classification}, in: \bibinfo{booktitle}{AISTATS}.
\bibitem[{Li et~al.(2016)Li, Guo and Schuurmans}]{2016Semi}
\bibinfo{author}{Li, X.}, \bibinfo{author}{Guo, Y.},
  \bibinfo{author}{Schuurmans, D.}, \bibinfo{year}{2016}.
\newblock \bibinfo{title}{Semi-supervised zero-shot classification with label
  representation learning}, in: \bibinfo{booktitle}{IEEE International
  Conference on Computer Vision}.
\bibitem[{Li et~al.(2018)Li, Zhang, Zhang and Huang}]{Li2018Discriminative}
\bibinfo{author}{Li, Y.}, \bibinfo{author}{Zhang, J.}, \bibinfo{author}{Zhang,
  J.}, \bibinfo{author}{Huang, K.}, \bibinfo{year}{2018}.
\newblock \bibinfo{title}{Discriminative learning of latent features for
  zero-shot recognition} .
\bibitem[{Long et~al.(2017)Long, Zhang, Xiao, Wei and Chang}]{Long2017Zero}
\bibinfo{author}{Long, C.}, \bibinfo{author}{Zhang, H.}, \bibinfo{author}{Xiao,
  J.}, \bibinfo{author}{Wei, L.}, \bibinfo{author}{Chang, S.F.},
  \bibinfo{year}{2017}.
\newblock \bibinfo{title}{Zero-shot visual recognition using
  semantics-preserving adversarial embedding network} .
\bibitem[{Meng and Guo(2018)}]{Meng2018Self}
\bibinfo{author}{Meng, Y.}, \bibinfo{author}{Guo, Y.}, \bibinfo{year}{2018}.
\newblock \bibinfo{title}{Self-training ensemble networks for zero-shot image
  recognition} .
\bibitem[{Mikolov et~al.(2014)Mikolov, Bengio, Singer, Shlens, Frome, Corrado
  and Dean}]{ZslCCSEICLR}
\bibinfo{author}{Mikolov, T.}, \bibinfo{author}{Bengio, S.},
  \bibinfo{author}{Singer, Y.}, \bibinfo{author}{Shlens, J.},
  \bibinfo{author}{Frome, A.}, \bibinfo{author}{Corrado, G.},
  \bibinfo{author}{Dean, J.}, \bibinfo{year}{2014}.
\newblock \bibinfo{title}{Zero-shot learning by convex combination of semantic
  embeddings}, in: \bibinfo{booktitle}{ICLR}.
\bibitem[{Pambala et~al.(2019)Pambala, Dutta and Biswas}]{Pambala2019Unified}
\bibinfo{author}{Pambala, A.K.}, \bibinfo{author}{Dutta, T.},
  \bibinfo{author}{Biswas, S.}, \bibinfo{year}{2019}.
\newblock \bibinfo{title}{Unified generator-classifier for efficient zero-shot
  learning} .
\bibitem[{Porter(2013)}]{Porter2013An}
\bibinfo{author}{Porter, M.F.}, \bibinfo{year}{2013}.
\newblock \bibinfo{title}{An algorithm for suffix stripping}, in:
  \bibinfo{booktitle}{Readings in Information Retrieval}.
\bibitem[{Qiao et~al.(2016)Qiao, Liu, Shen and Hengel}]{Qiao2016Less}
\bibinfo{author}{Qiao, R.}, \bibinfo{author}{Liu, L.}, \bibinfo{author}{Shen,
  C.}, \bibinfo{author}{Hengel, A.V.D.}, \bibinfo{year}{2016}.
\newblock \bibinfo{title}{Less is more: zero-shot learning from online textual
  documents with noise suppression} .
\bibitem[{Romera-Paredes and Torr(2015a)}]{romera2015embarrassingly}
\bibinfo{author}{Romera-Paredes, B.}, \bibinfo{author}{Torr, P.},
  \bibinfo{year}{2015}a.
\newblock \bibinfo{title}{An embarrassingly simple approach to zero-shot
  learning}, in: \bibinfo{booktitle}{International Conference on Machine
  Learning}, pp. \bibinfo{pages}{2152--2161}.
\bibitem[{Romera-Paredes and Torr(2015b)}]{Romera2015An}
\bibinfo{author}{Romera-Paredes, B.}, \bibinfo{author}{Torr, P.H.S.},
  \bibinfo{year}{2015}b.
\newblock \bibinfo{title}{An embarrassingly simple approach to zero-shot
  learning}, in: \bibinfo{booktitle}{Proceedings of the 32nd international
  conference on Machine learning (ICML '15)}.
\bibitem[{Sariyildiz and Cinbis(2019)}]{Sariyildiz_2019_CVPR}
\bibinfo{author}{Sariyildiz, M.B.}, \bibinfo{author}{Cinbis, R.G.},
  \bibinfo{year}{2019}.
\newblock \bibinfo{title}{Gradient matching generative networks for zero-shot
  learning}, in: \bibinfo{booktitle}{The IEEE Conference on Computer Vision and
  Pattern Recognition (CVPR)}.
\bibitem[{Shigeto et~al.(2015)Shigeto, Suzuki, Hara, Shimbo and
  Matsumoto}]{shigeto2015ridge}
\bibinfo{author}{Shigeto, Y.}, \bibinfo{author}{Suzuki, I.},
  \bibinfo{author}{Hara, K.}, \bibinfo{author}{Shimbo, M.},
  \bibinfo{author}{Matsumoto, Y.}, \bibinfo{year}{2015}.
\newblock \bibinfo{title}{Ridge regression, hubness, and zero-shot learning},
  in: \bibinfo{booktitle}{Joint European Conference on Machine Learning and
  Knowledge Discovery in Databases}, \bibinfo{organization}{Springer}. pp.
  \bibinfo{pages}{135--151}.
\bibitem[{Socher et~al.(2013)Socher, Ganjoo, Sridhar, Bastani, Manning and
  Ng}]{Socher2013Zero}
\bibinfo{author}{Socher, R.}, \bibinfo{author}{Ganjoo, M.},
  \bibinfo{author}{Sridhar, H.}, \bibinfo{author}{Bastani, O.},
  \bibinfo{author}{Manning, C.D.}, \bibinfo{author}{Ng, A.Y.},
  \bibinfo{year}{2013}.
\newblock \bibinfo{title}{Zero-shot learning through cross-modal transfer}, in:
  \bibinfo{booktitle}{International Conference on Neural Information Processing
  Systems}.
\bibitem[{{Suzuki} et~al.(2014){Suzuki}, {Sato}, {Oyama} and
  {Kurihara}}]{6974493}
\bibinfo{author}{{Suzuki}, M.}, \bibinfo{author}{{Sato}, H.},
  \bibinfo{author}{{Oyama}, S.}, \bibinfo{author}{{Kurihara}, M.},
  \bibinfo{year}{2014}.
\newblock \bibinfo{title}{Transfer learning based on the observation
  probability of each attribute}, in: \bibinfo{booktitle}{2014 IEEE
  International Conference on Systems, Man, and Cybernetics (SMC)}, pp.
  \bibinfo{pages}{3627--3631}.
\newblock \DOIprefix\doi{10.1109/SMC.2014.6974493}.
\bibitem[{Tao et~al.(2017)Tao, Zhang, Huang, Han and He}]{Tao2017AttnGAN}
\bibinfo{author}{Tao, X.}, \bibinfo{author}{Zhang, P.}, \bibinfo{author}{Huang,
  Q.}, \bibinfo{author}{Han, Z.}, \bibinfo{author}{He, X.},
  \bibinfo{year}{2017}.
\newblock \bibinfo{title}{Attngan: Fine-grained text to image generation with
  attentional generative adversarial networks} .
\bibitem[{Verma et~al.(2018)Verma, Arora, Mishra and Rai}]{2018Generalized}
\bibinfo{author}{Verma, V.K.}, \bibinfo{author}{Arora, G.},
  \bibinfo{author}{Mishra, A.}, \bibinfo{author}{Rai, P.},
  \bibinfo{year}{2018}.
\newblock \bibinfo{title}{Generalized zero-shot learning via synthesized
  examples}, in: \bibinfo{booktitle}{2018 IEEE/CVF Conference on Computer
  Vision and Pattern Recognition}.
\bibitem[{Wang et~al.(2017)Wang, Pu, Verma, Fan and Carin}]{2017Zero}
\bibinfo{author}{Wang, W.}, \bibinfo{author}{Pu, Y.}, \bibinfo{author}{Verma,
  V.K.}, \bibinfo{author}{Fan, K.}, \bibinfo{author}{Carin, L.},
  \bibinfo{year}{2017}.
\newblock \bibinfo{title}{Zero-shot learning via class-conditioned deep
  generative models}, in: \bibinfo{booktitle}{AAAI, 2018}.
\bibitem[{Wang et~al.(2018)Wang, Ye and Gupta}]{wang2018zero}
\bibinfo{author}{Wang, X.}, \bibinfo{author}{Ye, Y.}, \bibinfo{author}{Gupta,
  A.}, \bibinfo{year}{2018}.
\newblock \bibinfo{title}{Zero-shot recognition via semantic embeddings and
  knowledge graphs}, in: \bibinfo{booktitle}{Proceedings of the IEEE conference
  on computer vision and pattern recognition}, pp. \bibinfo{pages}{6857--6866}.
\bibitem[{Xian et~al.(2016)Xian, Akata, Sharma, Nguyen, Hein and
  Schiele}]{Xian_2016_CVPR}
\bibinfo{author}{Xian, Y.}, \bibinfo{author}{Akata, Z.},
  \bibinfo{author}{Sharma, G.}, \bibinfo{author}{Nguyen, Q.},
  \bibinfo{author}{Hein, M.}, \bibinfo{author}{Schiele, B.},
  \bibinfo{year}{2016}.
\newblock \bibinfo{title}{Latent embeddings for zero-shot classification}, in:
  \bibinfo{booktitle}{The IEEE Conference on Computer Vision and Pattern
  Recognition (CVPR)}.
\bibitem[{Xian et~al.(2018)Xian, Lorenz, Schiele and Akata}]{2018Feature}
\bibinfo{author}{Xian, Y.}, \bibinfo{author}{Lorenz, T.},
  \bibinfo{author}{Schiele, B.}, \bibinfo{author}{Akata, Z.},
  \bibinfo{year}{2018}.
\newblock \bibinfo{title}{Feature generating networks for zero-shot learning},
  in: \bibinfo{booktitle}{2018 IEEE/CVF Conference on Computer Vision and
  Pattern Recognition}.
\bibitem[{Xian et~al.(2017)Xian, Schiele and Akata}]{Xian2017Zero}
\bibinfo{author}{Xian, Y.}, \bibinfo{author}{Schiele, B.},
  \bibinfo{author}{Akata, Z.}, \bibinfo{year}{2017}.
\newblock \bibinfo{title}{Zero-shot learning - the good, the bad and the ugly}.
\bibitem[{Yang and Hospedales(2014)}]{yang2014unified}
\bibinfo{author}{Yang, Y.}, \bibinfo{author}{Hospedales, T.M.},
  \bibinfo{year}{2014}.
\newblock \bibinfo{title}{A unified perspective on multi-domain and multi-task
  learning}.
\newblock \bibinfo{journal}{arXiv preprint arXiv:1412.7489} .
\bibitem[{Yu and Aloimonos(2010)}]{10.1007/978-3-642-15555-0_10}
\bibinfo{author}{Yu, X.}, \bibinfo{author}{Aloimonos, Y.},
  \bibinfo{year}{2010}.
\newblock \bibinfo{title}{Attribute-based transfer learning for object
  categorization with zero/one training example}, in:
  \bibinfo{editor}{Daniilidis, K.}, \bibinfo{editor}{Maragos, P.},
  \bibinfo{editor}{Paragios, N.} (Eds.), \bibinfo{booktitle}{Computer Vision --
  ECCV 2010}, \bibinfo{publisher}{Springer Berlin Heidelberg},
  \bibinfo{address}{Berlin, Heidelberg}. pp. \bibinfo{pages}{127--140}.
\bibitem[{Zhang et~al.(2017)Zhang, Xiang and Gong}]{zhang2017learning}
\bibinfo{author}{Zhang, L.}, \bibinfo{author}{Xiang, T.},
  \bibinfo{author}{Gong, S.}, \bibinfo{year}{2017}.
\newblock \bibinfo{title}{Learning a deep embedding model for zero-shot
  learning}, in: \bibinfo{booktitle}{Proceedings of the IEEE Conference on
  Computer Vision and Pattern Recognition}, pp. \bibinfo{pages}{2021--2030}.
\bibitem[{Zhang and Saligrama(2015)}]{Zhang_2015_ICCV}
\bibinfo{author}{Zhang, Z.}, \bibinfo{author}{Saligrama, V.},
  \bibinfo{year}{2015}.
\newblock \bibinfo{title}{Zero-shot learning via semantic similarity
  embedding}, in: \bibinfo{booktitle}{The IEEE International Conference on
  Computer Vision (ICCV)}.
\bibitem[{Zhu et~al.(2018)Zhu, Elhoseiny, Liu, Peng and
  Elgammal}]{Zhu_2018_CVPR}
\bibinfo{author}{Zhu, Y.}, \bibinfo{author}{Elhoseiny, M.},
  \bibinfo{author}{Liu, B.}, \bibinfo{author}{Peng, X.},
  \bibinfo{author}{Elgammal, A.}, \bibinfo{year}{2018}.
\newblock \bibinfo{title}{A generative adversarial approach for zero-shot
  learning from noisy texts}, in: \bibinfo{booktitle}{The IEEE Conference on
  Computer Vision and Pattern Recognition (CVPR)}.

\end{thebibliography}

\end{document}